\documentclass[sigconf, nonacm]{acmart}

\AtBeginDocument{%
  }

\setcopyright{none}
\usepackage{booktabs}
\usepackage{multirow}
\usepackage{tabularx}

\usepackage[table]{xcolor}

\begin{document}

\title{DARE-EEG: A Foundation Model for Mining Dual-Aligned Representation of EEG}

\author{Yang Shao}
\orcid{0009-0005-8024-2521}
\affiliation{%
  \institution{College of Artificial Intelligence, Nanjing University of Aeronautics and Astronautics}
  \city{Nanjing}
  \state{Jiangsu}
  \country{China}
}
\email{shaoyang@nuaa.edu.cn}

\author{Peiliang Gong}
\orcid{0000-0003-2611-3145}
\affiliation{%
  \institution{College of Artificial Intelligence, Nanjing University of Aeronautics and Astronautics}
  \city{Nanjing}
  \state{Jiangsu}
  \country{China}
}
\email{plgong@nuaa.edu.cn}

\author{Qun Dai}
\orcid{0000-0003-4618-7299}
\affiliation{%
  \institution{College of Artificial Intelligence, Nanjing University of Aeronautics and Astronautics}
  \city{Nanjing}
  \state{Jiangsu}
  \country{China}
}
\email{daiqun@nuaa.edu.cn}

\author{Daoqiang Zhang}
\authornote{Corresponding author.}
\orcid{0000-0002-5658-7643}
\affiliation{%
  \institution{College of Artificial Intelligence, Nanjing University of Aeronautics and Astronautics}
  \city{Nanjing}
  \state{Jiangsu}
  \country{China}
}
\email{dqzhangn@nuaa.edu.cn}

\renewcommand{\shortauthors}{Shao et al.}
%%
%% Article type: Research, Review, Discussion, Invited or position
\acmArticleType{Research}
%%
%% Links to code and data
\acmCodeLink{https://github.com/borisveytsman/acmart}
\acmDataLink{htps://zenodo.org/link}
%%
%% Authors' contribution
\acmContributions{BT and GKMT designed the study; LT, VB, and AP
  conducted the experiments, BR, HC, CP and JS analyzed the results,
  JPK developed analytical predictions, all authors participated in
  writing the manuscript.}
%%
%% Sometimes the addresses are too long to fit on the page.  In this
%% case uncomment the lines below and fill them accodingly.
%%
%% \authorsaddresses{Corresponding author: Ben Trovato,
%% \href{mailto:trovato@corporation.com}{trovato@corporation.com};
%% Institute for Clarity in Documentation, P.O. Box 1212, Dublin,
%% Ohio, USA, 43017-6221}

\begin{abstract}
Foundation models pre-trained through masked reconstruction on large-scale EEG data have emerged as a promising paradigm for learning generalizable neural representations across diverse brain-computer interface applications. However, a critical yet overlooked challenge is that EEG encoders must learn representations invariant to incomplete observations—when different masked views of the same signal have minimal overlap, existing methods fail to constrain them to a consistent latent subspace, leading to degraded transferability. To address this, we propose DARE-EEG, a self-supervised foundation model that explicitly enforces the mask-invariance property through dual-aligned representation learning during pre-training. Specifically, we introduce mask alignment that constrains representations from multiple masked views of the same EEG sample via contrastive learning, complementing anchor alignment that aligns masked representations to momentum-updated complete features for semantic stability. Additionally, we propose conv-linear-probing, a parameter-efficient strategy that adapts pre-trained representations to heterogeneous electrode configurations and sampling rates through decoupled spectro-spatial projections. Extensive experiments across diverse EEG benchmarks demonstrate that DARE-EEG consistently achieves state-of-the-art in accuracy performance while maintaining relatively low parameter complexity and superior cross-dataset portability compared to existing methods. Furthermore, DARE-EEG contributes to effectively discovering and utilizing the rich potential representations in EEG.


% Electroencephalography (EEG) is one of the most widely used physiological signals for characterizing human cognitive states, containing rich yet underexplored neural representations. However, the inherently low signal-to-noise ratio of EEG, together with its variability across subjects, sessions, and tasks, poses significant challenges for learning transferable representations. In this paper, we propose DARE-EEG, a self-supervised framework designed to learn a universally applicable EEG encoder via masked reconstruction. DARE-EEG is built upon enforcing the mask-invariance property of EEG representations in the encoding space, combining mask alignment and anchor alignment to extract dual-aligned representations from EEG. At its core, DARE-EEG adopts a masked reconstruction paradigm consisting of an Encoder, a Predictor, and a Reconstructor. Anchor alignment is realized through a Target Encoder that provides stable semantic references, while mask alignment is enforced via contrastive learning across multiple masked views. Furthermore, we introduce an enhanced linear probing strategy to accommodate differences in electrode configurations and sampling rates across downstream tasks. Experiments on diverse EEG benchmarks demonstrate that DARE-EEG consistently achieves state-of-the-art or competitive performance with low parameter complexity and strong cross-dataset portability. DARE-EEG contributes to effectively discovering and utilizing the rich potential representations in EEG. The code is available at \url{https://anonymous.4open.science/r/DARE-EEG-72F7}.
\end{abstract}

\maketitle

\section{Introduction}
%%%%%
Electroencephalography (EEG) provides a non-invasive window into human brain activity, capturing rich neural dynamics that reflect cognitive states, emotional responses, and neurological conditions~\cite{ref1, ref2, ref3}. Its high temporal resolution, portability, and cost-effectiveness have made EEG indispensable across diverse applications, from clinical diagnosis~\cite{ref7, ref8} to brain-computer interfaces~\cite{ref9, ref10} and cognitive neuroscience~\cite{ref4}. However, learning robust and generalizable representations from EEG remains a fundamental challenge. The inherently low Signal-to-Noise Ratio (SNR) caused by electrode-skin impedance and volume conduction~\cite{ref11}, combined with substantial inter-subject variability and task-dependent fluctuations~\cite{ref12}, creates significant barriers to developing universally applicable EEG models.

\begin{figure}[t]
  \centering
  \includegraphics[width=0.9\columnwidth]{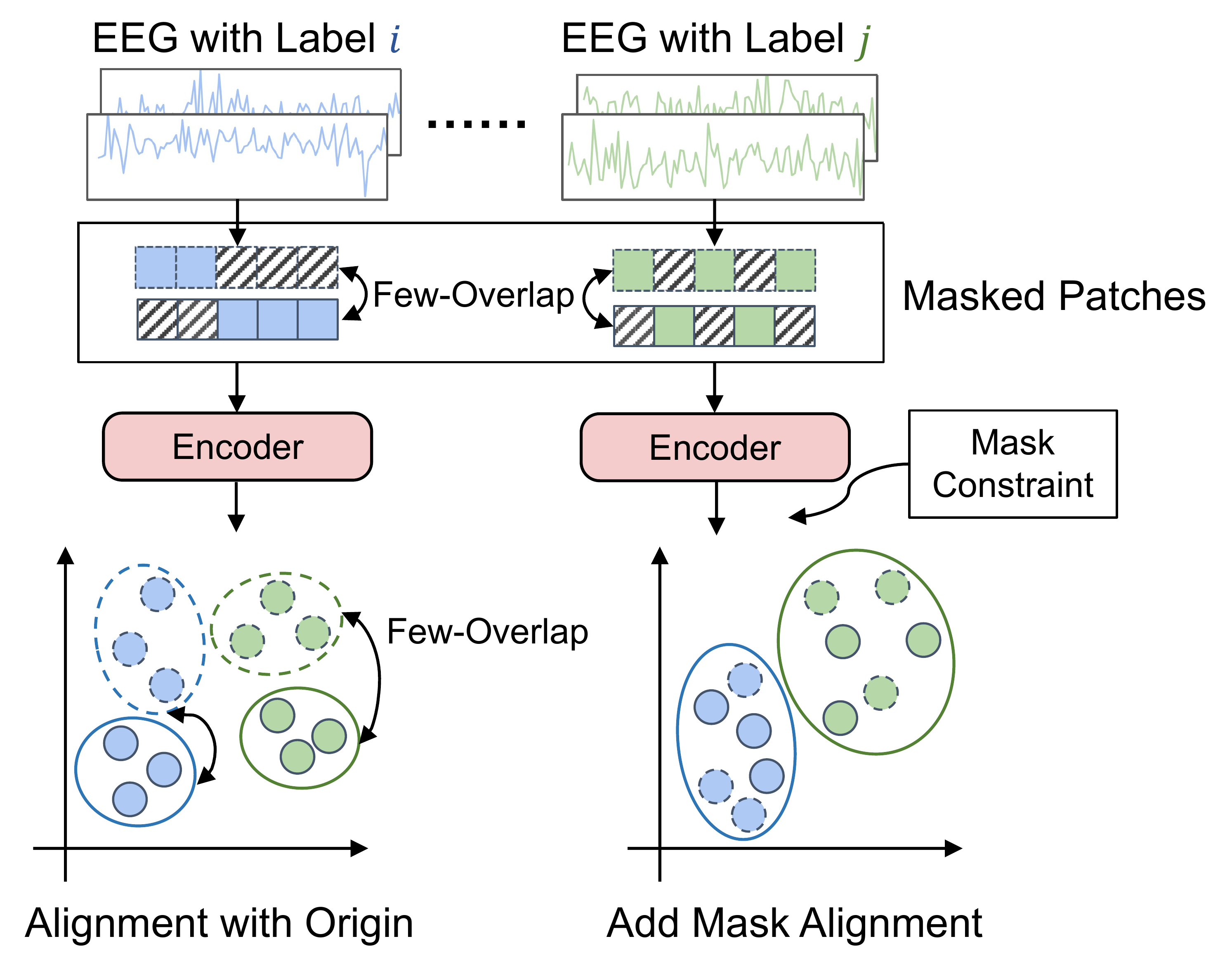}
  \caption{Illustration of the advantage of the dual alignment in DARE-EEG. Mask constraint attempts to preserve the invariant properties of the masked representation of EEG in the encoding space.}
  \label{fig1:intriduction}
\end{figure}

Traditional approaches addressing cross-subject and cross-task variability have primarily relied on domain adaptation techniques that explicitly align predefined source and target distributions~\cite{ref12, ref13}. While effective in controlled scenarios, these methods struggle to generalize beyond the specific domains they are designed to align, limiting their applicability to the diverse and heterogeneous nature of real-world EEG data. In contrast, the recent success of self-supervised learning in computer vision~\cite{ref15} and time-series analysis~\cite{ref14} has opened new avenues for learning transferable representations without explicit domain alignment. By pre-training on large-scale unlabeled data, self-supervised models can discover domain-invariant patterns that generalize across subjects, tasks, and recording conditions~\cite{ref16}.

Among self-supervised approaches, masked autoencoding has emerged as a particularly promising paradigm for EEG foundation models~\cite{ref17}. This approach trains encoders to reconstruct randomly masked portions of the input, encouraging the model to capture underlying structural dependencies rather than superficial patterns. Building on this principle, several EEG foundation models have been proposed. BENDR~\cite{ref18} uses a Convolutional Neural Network (CNN) to encode EEG sequences into BENDR vectors, applies random masking to the BENDR vectors, and then feeds them to a Transformer for reconstruction. BIOT~\cite{ref16} introduces a unified sentence structure to handle heterogeneous electrode configurations across datasets. LaBraM~\cite{ref19}, inspired by advances in large language models, segments EEG into channel-wise patches and employs vector quantization to create discrete neural codes. EEGMamba~\cite{ref20} replaces the Transformer backbone with Mamba for improved computational efficiency. More recently, EEGPT~\cite{ref22} incorporates Anchor Alignment (AA)—a technique from Data2Vec~\cite{ref21}—where masked representations are aligned to complete signal features from a momentum-updated target encoder, achieving state-of-the-art performance through linear probing.

Despite this progress, existing masked reconstruction approaches face a critical limitation when applied to EEG signals. Due to EEG's inherently low SNR and sensitivity to incomplete observations~\cite{ref24}, minor gaps or perturbations can substantially distort learned representations. This makes EEG particularly sensitive to incomplete observations compared to vision signals in reconstruction-based representation learning. Current methods employing AA ensure that masked views align with the complete signal but do not explicitly enforce consistency among different masked views of the same sample. This becomes particularly problematic when masked views have minimal overlap—a common scenario in high-masking-ratio pre-training. Without explicit constraints, the encoder may learn to encode mask patterns themselves rather than the underlying neural states, leading to what we term a violation of the \textit{Mask-Invariance Property} (MIP). MIP requires that representations derived from different masked views of the same EEG sample should be constrained to a consistent latent subspace while maintaining discriminability across different neural states. Existing models such as LaBraM and EEGPT, despite using channel position embeddings and AA, do not explicitly enforce this property, potentially limiting their ability to learn truly robust and transferable representations.

To address this fundamental limitation, we propose DARE-EEG (\textbf{D}ual-\textbf{A}ligned \textbf{RE}presentation learning for \textbf{EEG}), a foundation model that explicitly enforces mask invariance through a novel dual-alignment framework. As illustrated in Fig.~\ref{fig1:intriduction}, DARE-EEG integrates two complementary alignment mechanisms during masked reconstruction pre-training. First, AA aligns masked representations to stable semantic references from a momentum-updated target encoder, ensuring consistency with the complete signal. Second, and critically, Mask Alignment (MA) explicitly constrains representations from multiple masked views of the same EEG sample via contrastive learning, directly enforcing the MIP. This dual-alignment strategy prevents the encoder from degenerating into waveform memorization and instead encourages learning of meaningful neural state representations that are both mask-invariant and discriminative.
Beyond the core dual-alignment framework, we further introduce \textit{Conv-Linear-Probing} (CLP), an enhanced adaptation strategy for downstream tasks. Traditional linear probing~\cite{ref25} directly attaches a classification head to the frozen encoder, which can be suboptimal when downstream datasets have different electrode configurations or sampling rates than the pre-training data. CLP addresses this by inserting a lightweight spectro-spatial projection module before the encoder. This module employs decoupled convolutional operations to adaptively transform both the spatial (channel) and temporal (frequency) dimensions with minimal parameter overhead, enabling seamless transfer across heterogeneous EEG datasets.

Notably, we pre-trained DARE-EEG on over 264,000 EEG samples (over 917+ hours) spanning five diverse paradigms: emotion recognition (SEED~\cite{ref26}), motor imagery and execution (PhysioMI~\cite{ref27}), multi-paradigm cognitive tasks (M3CV~\cite{ref28}), steady-state visual evoked potentials (TSU~\cite{ref29}), and cognitive workload assessment (pBCIW~\cite{ref37}). The resulting Deep architecture contains 103 million parameters. We evaluated DARE-EEG with CLP on seven downstream benchmarks strictly separated from pre-training data. Results demonstrate that DARE-EEG performs exceptionally well in terms of Balanced Accuracy and is competitive overall across other metrics, 
% with improvements in balanced accuracy of up to 4.9\%, 
while not introducing an excessive increase in parameters compared with existing methods. The main contributions of this paper are:
\begin{itemize}
\item We identify the MIP as a critical yet overlooked requirement for learning robust EEG representations via masked reconstruction, and provide intuitive theoretical analysis (\textbf{\textit{Appendix~\ref{asec:mip_proof}}} illustrating its effectiveness and necessity).
\item We propose DARE-EEG, a foundation model that enforces MIP through dual-aligned representation learning, combining anchor alignment for semantic stability with mask alignment for explicit mask-invariance constraints via contrastive learning across multiple masked views.
\item We introduce conv-linear-probing, a parameter-efficient adaptation strategy that enables seamless transfer across heterogeneous EEG datasets with varying electrode configurations and sampling rates through decoupled spectro-spatial projections.
\item Through extensive experiments on seven diverse benchmarks, we demonstrate that DARE-EEG achieves competitive performance while maintaining relatively low parameter complexity, validating the effectiveness of explicit mask-invariance enforcement for transferable EEG representation learning.
\end{itemize}

\section{Methods}
\begin{figure*}[t]
  \centering
  \includegraphics[width=0.9\textwidth]{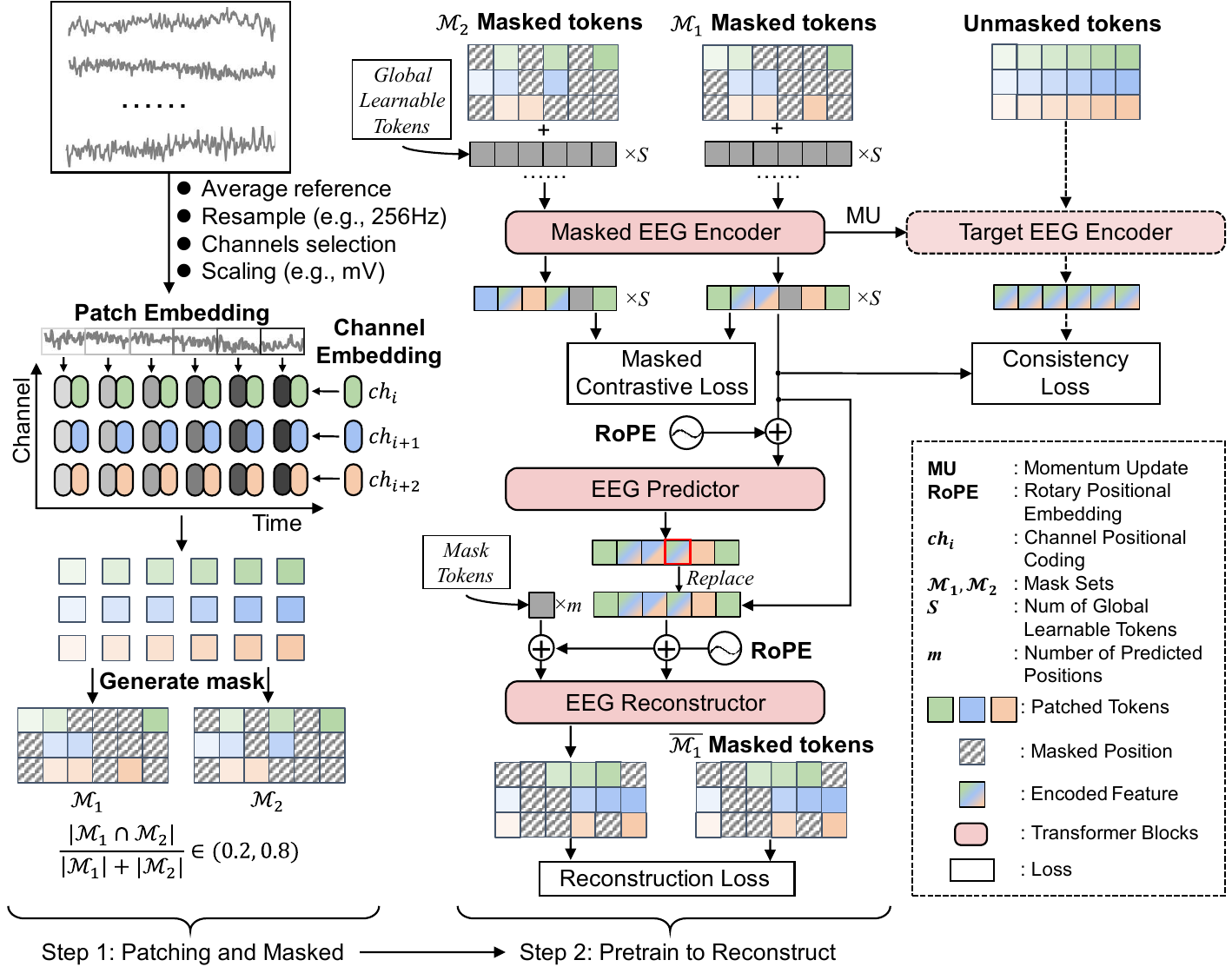}
  \caption{The structure of DARE-EEG used to discover dual-aligned representations of EEG signals. It mainly includes two steps: patching and masked, and pre-training to reconstruct EEG patches. The preprocessed unified EEG first undergoes a patching operation, followed by channel embedding. Then, two different masks are generated. The EEG patches affected by mask \(\mathcal{M}_{1}\) are processed through an encoder, predictor, and reconstructor for reconstruction, while simultaneously being aligned with the representations of the complete patches. The patches affected by mask \(\mathcal{M}_{2}\) are used for mask-alignment with the patches affected by \(\mathcal{M}_{1}\).}
  \label{fig2:model_structure}
\end{figure*}

Our method, DARE-EEG, is designed to constrain the EEG masked reconstruction process by jointly enforcing AA and MA during pre-training. The proposed framework consists of four key components as shown in Fig.~\ref{fig2:model_structure}: (1) Masked reconstruction based on a masked autoencoder~\cite{ref30}, used to learn a task-agnostic EEG encoder; (2) Anchor alignment via a consistency loss that encourages the encoded representations to preserve the original EEG features; (3) Mask alignment through explicit mask-aware constraints, ensuring the preservation of MIP between EEG representations in the encoded space; (4) An improved CLP based on linear-probing, which enhances the model's transferability to downstream tasks. The details of our method are described below.

\subsection{Masked Reconstruction}

\subsubsection{Masked autoencoder}

Masked autoencoders are designed to randomly mask a large proportion of input patches and reconstruct the missing content from the remaining visible context, thereby encouraging the encoder to learn meaningful and structured latent representations of the original input. Let $X_{in}\in\mathbb{R}^{C\times T}$ be the input data; the objective function for the masked reconstruction process is as follows:
\begin{equation}
\min_{\phi,\theta} \;
\frac{1}{\lvert 1 - \mathcal{M} \rvert}
\left\lVert
(1 - \mathcal{M}) \odot
\Bigl(
X_{\text{in}} - h_{\phi}\bigl(g_{\theta}(X_{\text{in}} \odot \mathcal{M})\bigr)
\Bigr)
\right\rVert.
\label{eq:1}
\end{equation}
Here, $\mathcal{M}$ represents the set of visible positions after masking, $\odot$ denotes the Hadamard product, $g$ and $h$ represent the encoder and decoder of the masked autoencoder, respectively, and $\theta$ and $\phi$ are their learnable parameters.

\subsubsection{Patching and Channel Embedding}

Unlike conventional regression tasks that aim to reconstruct the raw signal in a pointwise manner, DARE-EEG formulates masked reconstruction at the patch level, using patch-embedded EEG segments as the reconstruction targets. Patching not only significantly reduces the size of the input data, thereby improving training efficiency, but also, by aggregating local temporal context, helps the model focus on more stable and semantically meaningful neural activity patterns, rather than point-by-point fitting of high-frequency noise or transient fluctuations~\cite{ref31}. We assume that the patched $X_{in}$ becomes \(P=\{P_{ij}\}\), where \(i\) and \(j\) traverse the channel and time dimensions, respectively, with \(i\) ranging from 1 to \(C\) and \(j\) ranging from 1 to \(N\), where \(C\) is the total number of EEG electrode channels and \(N\) is the number of patches along the time dimension after patching. This means that the patching is performed in the temporal domain. Therefore, Eq.~\ref{eq:1} needs to be modified as follows to accommodate EEG patches:
\begin{equation}
\begin{aligned}
\min_{\phi,\theta} \;
&\frac{1}{\sum_{i,j} (1 - M_{i,j})}
\sum_{i,j}
(1 - M_{i,j})
\left\lVert
p_{i,j} - \hat{p}_{i,j}
\right\rVert, \\
&\hat{p}_{i,j} \in
h_{\phi}\!\left(g_{\theta}(\mathcal{M} \odot P)\right),
\end{aligned}
\label{eq:2}
\end{equation}
where \(M_{i,j}\in\mathcal{M}\), 1 indicates that the patch is visible to the encoder, while 0 indicates that it is invisible. Furthermore, DARE-EEG employs MSE as an evaluation metric in the reconstruction process; therefore, Eq.~\ref{eq:2} can be transformed into a reconstruction loss function \(\mathcal{L}_{RC}\):
\begin{equation}
    \mathcal{L}_{\mathrm{RC}}= \mathrm{MSE}\!\left((1-\mathcal{M}) \odot (P - \hat{P})\right),\quad \hat{P} = \{\hat{p}_{i,j}\}.
\label{eq:3}
\end{equation}

Moreover, some studies~\cite{ref16, ref22} have verified that channel embedding plays an important role in EEG foundation models. Incorporating channel embeddings not only facilitates the modeling of spatial heterogeneity across channels, but also improves the robustness of learned representations when adapting to datasets with varying channel configurations or electrode layouts. As shown in step 1 of Fig.~\ref{fig2:model_structure}, we use \({ch}_i\) to perform positional encoding for each channel. Specifically, we add the positional encoding to the EEG patches according to their corresponding channel positions, resulting in \(P_{CE} = {P_{11}+{ch}_1,P_{12}+{ch}_1,\ ...,P_{ij}+{ch}_i,...,P_{Cj}+{ch}_C}\). Thus, Eq.~\ref{eq:3} can be transformed into:
\begin{equation}
    \mathcal{L}_{\mathrm{RC}}= \mathrm{MSE}\!\left((1-\mathcal{M}) \odot (P_{CE} - \hat{P}_{CE})\right),
\label{eq:4}
\end{equation}
where \({\hat{P}}_{CE}\) is the output of the \(P_{CE}\) after the masking and reconstruction.

\subsubsection{Encoder, Predictor and Reconstructor}

To reduce model computational complexity, improve training efficiency, and stabilize the MIP, we introduced Global Learnable Tokens (GLTs) as a global semantic aggregator before the encoder. Unlike other models that use CLS tokens~\cite{ref32} and embed them in the temporal dimension, we embed GLTs in the channel space dimension to learn overall patterns across channels and time. GLTs are particularly important in EEG encoding because it helps align representations across different mask-views of the same EEG sample. The patches with \(S\) embedded GLTs become \(Tokens=Cat({\mathcal{M}\odot P}_{CE},\{{token}_1,...,\ {token}_S\}|dim=0)\). After passing through the encoder, we extract the GLTs separately and feed them into the predictor as \({Tokens}_{enc}\in \mathbb{R}^{S\times m N\times D}\), \(mN\) is the number of patches visible along the time dimension after masking:
\begin{equation}
    {Tokens}_{enc}=g_\theta\left(Tokens\right)\left[C+1:C+S\right].
\label{eq:5}
\end{equation}
The EEG predictor is responsible for predicting the masked portions, decoupling representation learning from the reconstruction objective~\cite{ref33}, thereby improving optimization stability. After being padded with masked positions, \textit{\({Tokens}_{enc}\)} underwent Rotational Positional Embedding (RoPE) so that the predictor could understand the temporal order of different positions. \({Tokens}_{pred}\in\mathbb{R}^{S\times N\times D}\) is shown below:
\begin{equation}
    {Tokens}_{pred}=g_\psi\left(\{t_{i,j}^{enc}+r_j\}\right),t_{i,j}^{enc}\in{Tokens}_{enc}.
\label{eq:6}
\end{equation}
Here, \(i\) ranges from 1 to \(S\), and \(j \) ranges from 1 to \(N\). \(r_{j}\) represents the RoPE coding. \(g_\psi\) represents the operation process of the predictor.

The EEG reconstructor does not directly receive the complete output of the predictor, as shown in Fig.~\ref{fig2:model_structure}, step 2, but only utilizes the predictor's output for the positions that were completely masked; for the remaining visible positions, it directly reuses the output representation of the encoder. This design preserves the integrity of the encoder representations while directing the predictor to focus exclusively on completing missing information, thus achieving a more favorable trade-off between reconstruction fidelity and representation generalization. Assume that \(t_{i,j}^{pred}\in{Tokens}_{pred}\), formal representation of \({Tokens}_{rec}\in\mathbb{R}^{C\times N}\) with RoPE is:
\begin{equation}
\begin{aligned}
\mathrm{Tokens}_{\mathrm{rec}}
=
h_{\phi}\!\Big(
\{ t^{\mathrm{enc}}_{i,j} + r_j \mid M_{i,j}=1 \}
\cup
\{ t^{\mathrm{pred}}_{i,j} + r_j \mid M_{i,j}=0 \}
\\[-2pt]
\cup
\{ t^{\mathrm{rec}}_{k} + r_k \}
\Big),
\end{aligned}
\label{eq:7}
\end{equation}
where \(t_k^{rec}\) represents the \(m\) positions that need to be reconstructed, where \(k\) ranges from 1 to \(m\). Thus, by substituting the output into Eq.~\ref{eq:4}, we can obtain the loss function for the mask reconstruction process of DARE-EEG:
\begin{equation}
    \mathcal{L}_{RC}=MSE\left(\left(1-\mathcal{M}\right)\odot(P_{CE}-{Tokens}_{rec})\right).
\label{eq:8}
\end{equation}

\subsection{Anchor Alignment}

Due to the challenges of low SNR and cross-domain shifts in EEG, a simple masked reconstruction process alone cannot effectively abstract stable EEG representations. Therefore, an additional target encoder is used to encode the unmasked EEG patches, serving as an anchor point with distinct original features. Target encoder ensures that the output of the EEG encoder is always aligned to a stable reference frame. Some studies~\cite{ref21, ref34} suggest that the target encoder can also mitigate the representational uncertainty caused by masked reconstruction. In this paper, the target encoder parameters \(\Delta\) is updated through Momentum Updates (MU)~\cite{ref35} from the EEG encoder. The update method is shown in Eq.~\ref{eq:9}:
\begin{equation}
    \Delta=\tau\Delta+\left(1-\tau\right)\theta,
\label{eq:9}
\end{equation}
where \(\tau=0.99\) denotes the momentum coefficient. The output of the target encoder with input \(P_{CE}\) can be represented as \({Tokens}_{tenc}\in\mathbb{R}^{S\times N \times D}\):
\begin{equation}
    {Tokens}_{tenc}=f_\Delta\left(Cat\left(P_{CE},GLTs\middle|dim=0\right)\right)\left[C+1:C+S\right].
\label{eq:10}
\end{equation}

By fixing \({Tokens}_{tenc}\) as an anchor, we use MSE as a consistency loss to constrain the EEG encoder for AA. The loss function is an instantiation of Eq.~\ref{eq:4}:
\begin{equation}
    \mathcal{L}_{AA}=MSE\left({Tokens}_{tenc},{Tokens}_{enc}\right).
\label{eq:11}
\end{equation}

\begin{figure*}[htpb]
  \centering
  \includegraphics[width=0.9\textwidth]{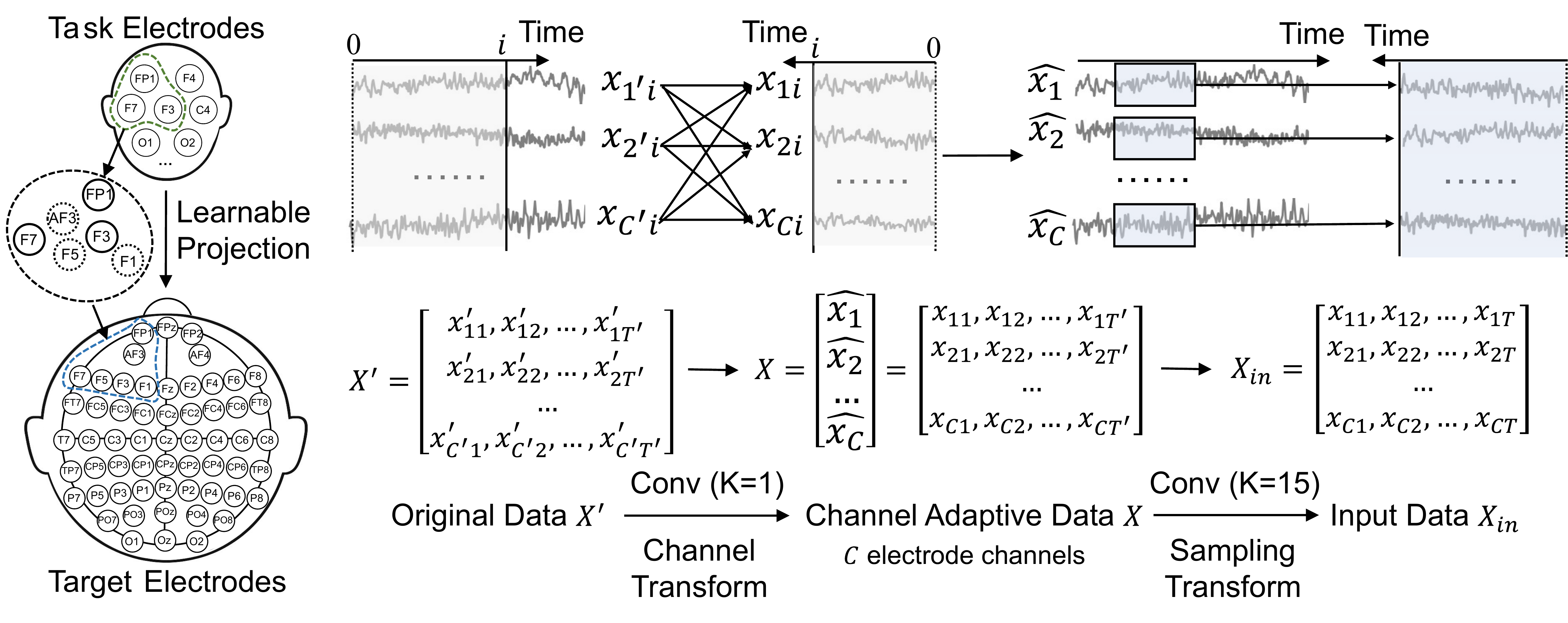}
  \caption{Conv-Linear-Probing learns an effective mapping method for different downstream tasks by decoupling channel adaptation from Sampling adaptation. Target Electrodes refers to the unified EEG electrode channels and format we used during pre-training. After Channel Transform, Task Electrodes, due to subsequent channel position embedding, are forced by the training loss to compensate for the desired channels.}
  \label{fig3:convlp}
\end{figure*}

\subsection{Mask Alignment}

MA is a critical mechanism for enforcing the MIP of EEG representations in the latent space. Specifically, for each embedded \(P_{CE}^i\) in a mini-batch, two distinct random masking patterns are generated, as illustrated in Fig.~\ref{fig2:model_structure} (denoted as \(\mathcal{M}_{1}\) and \(\mathcal{M}_{2}\)). The \(\mathcal{M}_{1}\) is applied during the masked reconstruction process, while the \(\mathcal{M}_{2}\) is exclusively used for MA, enabling explicit consistency constraints across different mask-views of the same EEG signal. To avoid degenerate mask configurations, we do not allow the two random masks to be either completely disjoint or excessively overlapping. Extremely low overlap may result in insufficient shared context for reliable alignment, whereas excessive overlap diminishes the diversity between masked views. Accordingly, the overlap ratio is constrained to the interval \((0.2, 0.8)\).

Secondly, \(\mathcal{M}_{1}\) and \(\mathcal{M}_{2}\) are processed by the same EEG encoder, yielding two sets of latent representations \(Tokens_{M1}\) and \(Tokens_{M2}\). Within a mini-batch, multiple pairs of masked views are therefore generated. MA encourages representations originating from different mask-views of the same EEG sample to be close in the latent space, while simultaneously pushing representations of different samples further apart. It should be noted that this constraint operates purely at the representation level and does not require the use of original labels; therefore, it remains self-supervised learning, and does not introduce label leakage. This process can be formally represented as follows:
\begin{equation}
\begin{aligned}
\mathrm{Tokens}^{i}_{\mathcal{M}_1}
&=
g_{\theta}\!\left(
\mathrm{Cat}\!\left(\mathcal{M}_1 \odot P^{i}_{\mathrm{CE}},\, \mathrm{GLTs} \right)
\right)[C+1:C+S], \\
\mathrm{Tokens}^{i}_{\mathcal{M}_2}
&=
g_{\theta}\!\left(
\mathrm{Cat}\!\left(\mathcal{M}_2 \odot P^{i}_{\mathrm{CE}},\, \mathrm{GLTs} \right)
\right)[C+1:C+S].
\end{aligned}
\label{eq:12}
\end{equation}
Here, \(i\) ranges from 1 to \(B\), where \(B\) is the size of a mini-batch. Third, the cosine similarity is calculated:
\begin{equation}
s_{i,j}=\frac{PP(\mathrm{Tokens}^{i}_{\mathcal{M}_1})\,
\mathrm{T}\!\left(PP(\mathrm{Tokens}^{j}_{\mathcal{M}_2})\right)}{\kappa},\quad
i,j \in \{1,2,\ldots,B\}.
\label{eq:13}
\end{equation}
Fourth, we use symmetric InfoNCE as the masked constraint loss \(\mathcal{L}_{MA}\), and by optimizing this loss, we can achieve MA:

\begin{equation}
\begin{aligned}
\mathcal{L}_{1\rightarrow 2}
&=-\frac{1}{B}\sum_{i=1}^{B}\log\frac{\exp(s_{i,i})}{\sum_{j=1}^{B} \exp(s_{i,j})}, \\
\mathcal{L}_{2\rightarrow 1}
&=-\frac{1}{B}\sum_{i=1}^{B}\log\frac{\exp(s_{i,i})}{\sum_{j=1}^{B} \exp(s_{j,i})}, \\
\mathcal{L}_{\mathrm{MA}}
&=\frac{1}{2}\left(\mathcal{L}_{1\rightarrow 2}+\mathcal{L}_{2\rightarrow 1}\right).
\end{aligned}
\label{eq:14}
\end{equation}
In \(\mathcal{L}_{MA}\), \(\kappa\) is a trainable parameter that will be automatically optimized during the training process, with an initial value set to 0.1. To improve stability during training, the masked representations were subjected to pooling and projection operations (\(PP\)) before calculating similarity.
Finally, the parameters of DARE-EEG during the pre-training are optimized by all three loss functions simultaneously, and its training loss function is \(\mathcal{L}\):
\begin{equation}
    \mathcal{L}=\mathcal{L}_{RC}+\mathcal{L}_{AA}+0.1\mathcal{L}_{MA}.
\label{eq:15}
\end{equation}

\subsection{Conv-linear-probing in Downstream Tasks}

\subsubsection{Conv-linear-probing}

\begin{figure}[]
  \centering
  \includegraphics[width=0.8\columnwidth]{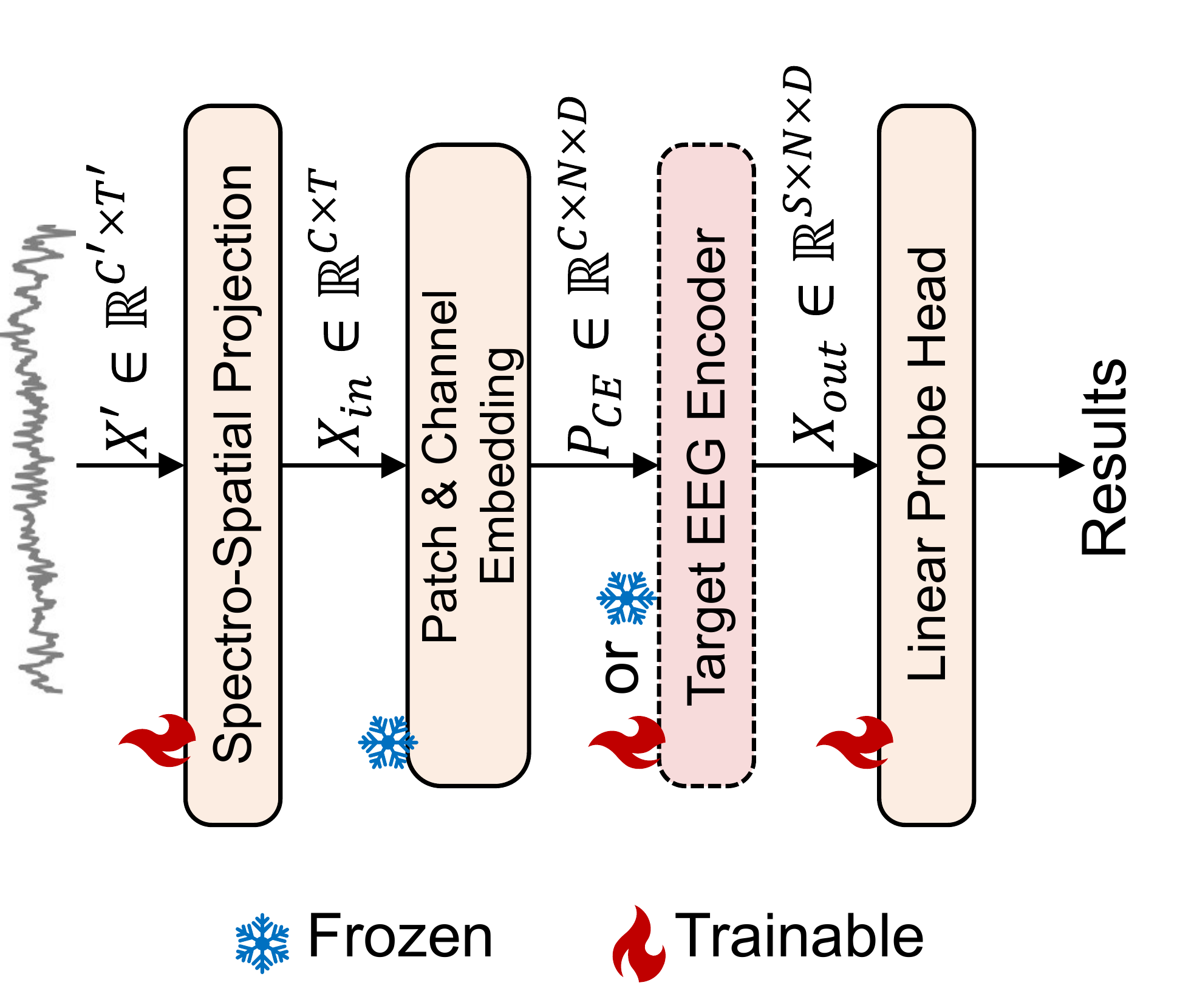}
  \caption{Overview of the DARE-EEG downstream training pipeline. The pre-trained encoder is employed as the core feature extractor, while the spectro-spatial projection and the linear probe head jointly constitute the CLP for supervised downstream tasks.}
  \label{fig4:downtasks}
\end{figure}

For downstream tasks, DARE-EEG employs the pre-trained encoder solely as an EEG feature extractor. To facilitate task-specific adaptation, a standard linear probe is attached to the encoder head to map the learned features to classification outputs. In addition, a lightweight Spectro–spatial projection module is introduced at the encoder tail to accommodate variations in channel configurations and sampling frequencies across datasets. The overall downstream adaptation pipeline is illustrated in Fig.~\ref{fig3:convlp}. Target Electrodes displays the unified channels used in pre-training. Specifically, the input EEG signal \(X’\) is first subjected to channel adaptation via a \textit{\(1\times1\)} convolution, which maps non-standard electrode configurations to a unified channel space compatible with channel embedding. Subsequently, a large-kernel convolution~\cite{ref36} is applied along the temporal dimension of each channel, enabling the model to capture long-range temporal dependencies and adapt to heterogeneous frequency characteristics. The result, \(X_{in}\), is the input to the patching operation.

\subsubsection{Supervised Training in Downstream Tasks}

Fig.~\ref{fig4:downtasks} shows the specific architecture of DARE-EEG equipped with the proposed CLP for supervised training on downstream tasks. In one approach, the pre-trained encoder is kept frozen and only the CLP is optimized; another approach is to allow the encoder to participate in fine-tuning, which is more suitable for large-scale datasets with high classification difficulty (i.e., TUH-EEG). Let the model's output be Outputs, both approaches are optimized using the cross-entropy loss:
\begin{equation}
\mathcal{L}_{\mathrm{CE}}
= - \log \frac{\exp(\mathrm{Output}_{l})}
{\sum_{y \in Y} \mathrm{Output}_{y}},
\quad Y \in \{0,1,\ldots,\mathrm{num}~{\mathrm{classes}}\}.
\label{eq:ce_loss}
\end{equation}
Here, \({\rm Output}_y\) is the logit corresponding to class \(y\), and \({\rm Output}_l\) is the logit of the true class \(l\). The proposed CLP effectively mitigates overfitting when fine-tuning large-scale models with limited labeled samples, while maintaining strong portability across different downstream tasks. Moreover, since the additional modules are simple and parameter-efficient, the downstream performance largely depends on the encoder, facilitating a more faithful assessment of the learned representations.

\section{Experiments and Results}

\subsection{Datasets and Implementation details}

\begin{table}[t]
\centering
\small
\caption{Summary of pretraining and downstream EEG datasets.}
\label{tab:1}
\renewcommand{\arraystretch}{0.92}
\begin{tabular}{l l l l}
\toprule
Dataset & Paradigm & Tasks & Subjects \\
\midrule
\multicolumn{4}{c}{\textbf{Pretraining Datasets}} \\
\midrule
SEED      & Video Stimulation & Emotion                         & 15    \\
PhysioMI & MI and ME          & Left vs Right                   & 109   \\
M3CV      & MSU, MSE and MPA   & Identification & 106  \\
TSU       & SSVEP              & Frequency Stimulation           & 33    \\
pBCIW     & MATB               & Cognitive Workload              & 15   \\
\midrule
\multicolumn{4}{c}{\textbf{Downstream Datasets}} \\
\midrule
TUH-AB   & Clinical EEG        & Normal vs Abnormal              & 2383  \\
TUH-EV   & Clinical EEG        & Event Detection                 & 288   \\
BCIC-2A  & MI                  & 4-class MI                      & 9     \\
BCIC-2B  & MI                  & Left vs Right                   & 9     \\
SEEDIV   & Video Stimulation   & Emotion (4 classes)             & 15    \\
SleepEDF & Polysomnography     & Sleep Staging      & 197  \\
MMWM     & Working Memory      & Cognitive Workload              & 30    \\
\bottomrule
\end{tabular}
\end{table}

\begin{table}[t]
\centering
\caption{Performance on TUAB for abnormal EEG detection (mean $\pm$ std).}
\label{tab2:tuab_results}
\small
\setlength{\tabcolsep}{4pt}
\begin{tabularx}{\columnwidth}{l c c c}
\toprule
Models & Balanced ACC & AUC\_PR & AUROC \\
\midrule
SPaRCNet~\cite{ref42:SpaRCNet} & $0.7896\!\pm\!0.0018$ & $0.8414\!\pm\!0.0018$ & $0.8676\!\pm\!0.0012$ \\
ContraWR~\cite{ref43:ContraWR} & $0.7746\!\pm\!0.0041$ & $0.8421\!\pm\!0.0104$ & $0.8456\!\pm\!0.0074$ \\
CNN-Trans~\cite{ref44:CNNT} & $0.7777\!\pm\!0.0022$ & $0.8433\!\pm\!0.0039$ & $0.8461\!\pm\!0.0013$ \\
FFCL~\cite{ref45:FFCL} & $0.7848\!\pm\!0.0038$ & $0.8448\!\pm\!0.0065$ & $0.8569\!\pm\!0.0051$ \\
ST-Trans~\cite{ref46:STT} & $0.7966\!\pm\!0.0023$ & $0.8521\!\pm\!0.0026$ & $0.8707\!\pm\!0.0019$ \\
BIOT (PREST)~\cite{ref16} & $0.7907\!\pm\!0.0050$ & $0.8752\!\pm\!0.0051$ & $0.8730\!\pm\!0.0021$ \\
BIOT (Other)~\cite{ref16} & $0.7959\!\pm\!0.0057$ & $0.8792\!\pm\!0.0023$ & $0.8815\!\pm\!0.0043$ \\
EEGPT-tiny~\cite{ref22} & $0.7959\!\pm\!0.0021$ & -- & $0.8716\!\pm\!0.0041$ \\
EEGPT-large~\cite{ref22} & $0.7983\!\pm\!0.0030$ & -- & $0.8718\!\pm\!0.0050$ \\
\midrule
\rowcolor{gray!10}
\textbf{DARE-EEG-Base} & $\mathbf{0.8145\!\pm\!0.0031}$ & $\mathbf{0.9138\!\pm\!0.0039}$ & $\mathbf{0.8927\!\pm\!0.0020}$ \\
\rowcolor{gray!20}
\textbf{DARE-EEG-Deep} & $\mathbf{0.8156\!\pm\!0.0022}$ & $\mathbf{0.9060\!\pm\!0.0041}$ & $\mathbf{0.8904\!\pm\!0.0010}$ \\
\bottomrule
\end{tabularx}
\end{table}

\begin{table}[t]
\centering
\caption{Performance on the TUEV dataset for event type classification (mean $\pm$ std).}
\label{tab3:tuev_results}
\small
\setlength{\tabcolsep}{4pt} 
\renewcommand{\arraystretch}{0.90}
\begin{tabularx}{\columnwidth}{l c c c}
\toprule
Models & Balanced ACC & Kappa & Weighted F1 \\
\midrule
SPaRCNet~\cite{ref42:SpaRCNet} & $0.4161\!\pm\!0.0262$ & $0.4233\!\pm\!0.0181$ & $0.7024\!\pm\!0.0104$ \\
ContraWR~\cite{ref43:ContraWR} & $0.4384\!\pm\!0.0349$ & $0.3912\!\pm\!0.0237$ & $0.6893\!\pm\!0.0136$ \\
CNN-Trans~\cite{ref44:CNNT} & $0.4087\!\pm\!0.0161$ & $0.3815\!\pm\!0.0134$ & $0.6854\!\pm\!0.0293$ \\
FFCL~\cite{ref45:FFCL} & $0.3979\!\pm\!0.0104$ & $0.3732\!\pm\!0.0188$ & $0.6783\!\pm\!0.0120$ \\
ST-Trans~\cite{ref46:STT} & $0.3984\!\pm\!0.0228$ & $0.3765\!\pm\!0.0306$ & $0.6823\!\pm\!0.0190$ \\
BIOT (PREST)~\cite{ref16} & $0.5207\!\pm\!0.0285$ & $0.4932\!\pm\!0.0301$ & $0.7381\!\pm\!0.0169$ \\
BIOT (Other)~\cite{ref16} & $0.5281\!\pm\!0.0225$ & $0.5273\!\pm\!0.0249$ & $0.7492\!\pm\!0.0082$ \\
EEGPT-tiny~\cite{ref22} & $0.5670\!\pm\!0.0066$ & $0.5085\!\pm\!0.0173$ & $0.7535\!\pm\!0.0097$ \\
EEGPT-large~\cite{ref22} & $0.6232\!\pm\!0.0114$ & $0.6351\!\pm\!0.0134$ & $0.8187\!\pm\!0.0063$ \\
\midrule
\rowcolor{gray!10}
\textbf{DARE-EEG-Base} & $\mathbf{0.6561\!\pm\!0.0157}$ & $\mathbf{0.5827\!\pm\!0.0283}$ & $\mathbf{0.7884\!\pm\!0.0161}$ \\
\rowcolor{gray!20}
\textbf{DARE-EEG-Deep} & $\mathbf{0.6489\!\pm\!0.0021}$ & $\mathbf{0.5883\!\pm\!0.0012}$ & $\mathbf{0.7920\!\pm\!0.0004}$ \\
\bottomrule
\end{tabularx}
\end{table}

\begin{table*}[t]
\centering
\caption{Performance comparison on downstream datasets. The optimal results are highlighted in bold. Pretrain indicates whether pre-trained weights are used, and T.P. represents the number of parameters involved in training in downtasks.}
\label{tab4:downstream_main}
\small
\setlength{\tabcolsep}{3.2pt}
\renewcommand{\arraystretch}{0.90}

\begin{tabularx}{0.9\textwidth}{%
c l c r
>{\centering\arraybackslash}X
>{\centering\arraybackslash}X
>{\centering\arraybackslash}X}
\toprule
\textbf{Dataset} & \textbf{Methods} & \textbf{Pretrain} & \textbf{T.P. (K)} &
\textbf{Balanced Accuracy} & \textbf{Cohen's Kappa} & \textbf{Weighted F1 / ROC\_AUC} \\
\midrule

% ================= BCIC-2A =================
\multirow{8}{*}{\textbf{BCIC-2A}}
& EEGNet     & $\times$ & 6.0   & 0.5153 $\pm$ 0.0374 & 0.3537 $\pm$ 0.0499 & 0.5044 $\pm$ 0.0366 \\
& DeepCNN    & $\times$ & 287.0 & 0.5312 $\pm$ 0.0306 & 0.3750 $\pm$ 0.0407 & 0.5053 $\pm$ 0.0381 \\
& Conformer  & $\times$ & 160.0 & 0.5206 $\pm$ 0.0165 & 0.3608 $\pm$ 0.0220 & 0.5043 $\pm$ 0.0167 \\
& BENDR      & $\checkmark$ & 22.5 & 0.4899 $\pm$ 0.0070 & 0.3199 $\pm$ 0.0094 & 0.4836 $\pm$ 0.0076 \\
& BIOT       & $\checkmark$ & 31.6 & 0.4590 $\pm$ 0.0196 & 0.2787 $\pm$ 0.0261 & 0.4282 $\pm$ 0.0289 \\
& LaBraM     & $\checkmark$ & 102.0& 0.5613 $\pm$ 0.0052 & 0.4151 $\pm$ 0.0069 & 0.5520 $\pm$ 0.0052 \\
& EEGPT      & $\checkmark$ & 34.2 & 0.5846 $\pm$ 0.0070 & 0.4462 $\pm$ 0.0094 & 0.5715 $\pm$ 0.0051 \\
\rowcolor{gray!20}
& Ours       & $\checkmark$ & 35.4 &
\textbf{0.5911 $\pm$ 0.0036} &
\textbf{0.4547 $\pm$ 0.0048} &
\textbf{0.5855 $\pm$ 0.0033} \\
\midrule

% ================= BCIC-2B =================
\multirow{8}{*}{\textbf{BCIC-2B}}
& EEGNet     & $\times$ & 5.2   & 0.7179 $\pm$ 0.0202 & 0.4357 $\pm$ 0.0405 & 0.7845 $\pm$ 0.0275 \\
& DeepCNN    & $\times$ & 270.0 & 0.7178 $\pm$ 0.0028 & 0.4357 $\pm$ 0.0057 & {0.8010 $\pm$ 0.0015} \\
& Conformer  & $\times$ & 127.0 & 0.6594 $\pm$ 0.0051 & 0.3187 $\pm$ 0.0103 & 0.7379 $\pm$ 0.0074 \\
& BENDR      & $\checkmark$ & 22.7 & 0.7067 $\pm$ 0.0011 & 0.4131 $\pm$ 0.0022 & 0.7854 $\pm$ 0.0029 \\
& BIOT       & $\checkmark$ & 1.1  & 0.6409 $\pm$ 0.0118 & 0.2817 $\pm$ 0.0236 & 0.7095 $\pm$ 0.0141 \\
& LaBraM     & $\checkmark$ & 66.2 & 0.6851 $\pm$ 0.0063 & 0.3703 $\pm$ 0.0125 & 0.7576 $\pm$ 0.0067 \\
& EEGPT      & $\checkmark$ & 33.4 & 0.7212 $\pm$ 0.0019 & 0.4426 $\pm$ 0.0037 & 0.8059 $\pm$ 0.0032 \\
\rowcolor{gray!20}
& Ours       & $\checkmark$ & 33.5 &
\textbf{0.7471 $\pm$ 0.0017} &
\textbf{0.4941 $\pm$ 0.0033} &
\textbf{0.8151 $\pm$ 0.0015} \\
\midrule

\multirow{8}{*}{\textbf{SEEDIV}} 
& EEGNet     & $\times$ &   8.2  & 0.3466 $\pm$ 0.0043 & 0.1398 $\pm$ 0.0062 & 0.3263 $\pm$ 0.0036 \\
& DeepCNN    & $\times$ & 312.0  & 0.4203 $\pm$ 0.0118 & 0.2341 $\pm$ 0.0175 & 0.3902 $\pm$ 0.0149 \\
& Conformer  & $\times$ & 221.0  & 0.4286 $\pm$ 0.0050 & 0.2453 $\pm$ 0.0056 & 0.3977 $\pm$ 0.0103 \\
& BENDR      & $\checkmark$ &  22.5  & 0.3664 $\pm$ 0.0077 & 0.1624 $\pm$ 0.0113 & 0.3651 $\pm$ 0.0106 \\
& BIOT       & $\checkmark$ &  32.2  & 0.4173 $\pm$ 0.0041 & 0.2248 $\pm$ 0.0041 & 0.3714 $\pm$ 0.0092 \\
& LaBraM     & $\checkmark$ &  31.8  & 0.4244 $\pm$ 0.0109 & 0.2323 $\pm$ 0.0138 & 0.4060 $\pm$ 0.0116 \\
& EEGPT      & $\checkmark$ &  33.8  & 0.3920 $\pm$ 0.0033 & 0.1908 $\pm$ 0.0036 & 0.3934 $\pm$ 0.0090 \\
\rowcolor{gray!20}
& Ours       & $\checkmark$ &  36.8  & \textbf{0.4357 $\pm$ 0.0033} & \textbf{0.2454 $\pm$ 0.0030} & \textbf{0.4300 $\pm$ 0.0045} \\
\midrule

\multirow{8}{*}{\textbf{SleepEDF}} 
&EEGNet     & $\times$ &  15.2  & 0.6281 $\pm$ 0.0122 & 0.5752 $\pm$ 0.0075 & 0.6719 $\pm$ 0.0060 \\
&DeepCNN    & $\times$ & 301.0  & 0.6765 $\pm$ 0.0158 & 0.6614 $\pm$ 0.0088 & 0.7523 $\pm$ 0.0064 \\
&Conformer  & $\times$ & 125.0  & 0.6469 $\pm$ 0.0045 & 0.6010 $\pm$ 0.0051 & 0.6989 $\pm$ 0.0039 \\
&BENDR      & $\checkmark$ &  28.3  & 0.6968 $\pm$ 0.0031 & \textbf{0.6884 $\pm$ 0.0033} & \textbf{0.7581 $\pm$ 0.0032} \\
&BIOT       & $\checkmark$ &   1.3  & 0.6907 $\pm$ 0.0033 & 0.6570 $\pm$ 0.0082 & 0.7442 $\pm$ 0.0047 \\
&LaBraM     & $\checkmark$ & 127.4  & 0.6596 $\pm$ 0.0142 & 0.5985 $\pm$ 0.0009 & 0.6974 $\pm$ 0.0112 \\
&EEGPT      & $\checkmark$ & 398.3  & 0.6803 $\pm$ 0.0009 & 0.6634 $\pm$ 0.0082 & 0.7451 $\pm$ 0.0062 \\
\rowcolor{gray!20}
&Ours       & $\checkmark$ & 169.6  & \textbf{0.7011 $\pm$ 0.0016} & 0.6586 $\pm$ 0.0088 & 0.7220 $\pm$ 0.0267 \\
\midrule

\multirow{8}{*}{\textbf{MMWM}} 
&EEGNet     & $\times$ &   4.2  & 0.5660 $\pm$ 0.0131 & 0.1319 $\pm$ 0.0260 & 0.5727 $\pm$ 0.0164 \\
&DeepCNN    & $\times$ & 284.0  & 0.5645 $\pm$ 0.0202 & 0.1292 $\pm$ 0.0407 & 0.5953 $\pm$ 0.0495 \\
&Conformer  & $\times$ & 170.0  & 0.6131 $\pm$ 0.0306 & 0.2262 $\pm$ 0.0610 & 0.6573 $\pm$ 0.0420 \\
&BENDR      & $\checkmark$ & 196.1  & 0.6183 $\pm$ 0.0029 & 0.2365 $\pm$ 0.0058 & 0.6533 $\pm$ 0.0043 \\
&BIOT       & $\checkmark$ &  31.2  & 0.6253 $\pm$ 0.0058 & 0.2505 $\pm$ 0.0119 & 0.6667 $\pm$ 0.0079 \\
&LaBraM     & $\checkmark$ &  13.4  & 0.5965 $\pm$ 0.0067 & 0.1931 $\pm$ 0.0132 & 0.6159 $\pm$ 0.0090 \\
&EEGPT      & $\checkmark$ &  33.3  & 0.6038 $\pm$ 0.0022 & 0.2076 $\pm$ 0.0044 & 0.6320 $\pm$ 0.0034 \\
\rowcolor{gray!20}
&Ours       & $\checkmark$ &  35.4  & \textbf{0.6490 $\pm$ 0.0063} & \textbf{0.2976 $\pm$ 0.0124} & \textbf{0.6915 $\pm$ 0.0122} \\
\bottomrule
\end{tabularx}
\end{table*}

Table~\ref{tab:1} provides information on the datasets used by DARE-EEG in pre-training and downstream tasks. Detailed descriptions of datasets are provided in the \textbf{\textit{Appendix~\ref{asec:dataset}}}. The tasks involved in pre-training mainly include emotion recognition using the SEED~\cite{ref26} dataset, motor imagery and execution using the PhysioMI~\cite{ref27} dataset, multi-subject, multi-session, and multi-paradigm cognitive and visual tasks using the M3CV~\cite{ref28} dataset, steady-state visual evoked potentials using the TSU~\cite{ref29} dataset, and MATB cognitive workload assessment using the pBCIW~\cite{ref37} dataset. For downstream tasks, we selected seven commonly used datasets, including the clinical EEG datasets TUAB and TUEV from the TUH-EEG dataset~\cite{ref38:tuh}, motor imagery classification from the BCIC-2A and BCIC-2B \allowbreak datasets~\cite{ref39:bcic}, emotion recognition from the SEEDIV dataset~\cite{ref40:seediv}, sleep staging from SleepEDF~\cite{ref41:sleep}, and working memory cognitive load recognition from the MMWM dataset~\cite{ref23}.

During pre-training, to ensure training stability and prevent dataset-specific biases, all EEG recordings were uniformly downsampled to 256 Hz, and only overlapping channels across datasets were retained. The resulting channel configuration follows the international 10–20 system, and the selected electrodes are highlighted as target channels in Fig.~\ref{fig3:convlp}. Furthermore, to account for variations across EEG tasks, certain datasets were additionally filtered and interpolated before being uniformly segmented into fixed-length segments of 1024. The temporal patch length is fixed to 64 samples, resulting in \(N=16\) patches along the time dimension. During masking, each temporal patch is fully masked with a probability of 50\%. For the remaining patches, only 20\% of the channels are randomly retained as visible context, while the rest are masked. Pre-training was conducted using a OneCycle learning rate schedule with a maximum learning rate of 5e-4 and a division factor of 24. The batch size was set to 64, and the max epochs was set to 200. All pre-training data were randomly split into training and validation sets with a ratio of 9:1.

In downstream tasks, a Leave-One-Subject-Out (LOSO) cross-validation strategy was used on the BCIC-2A, BCIC-2B, and SEEDIV datasets, while a mixed training strategy was used on the SleepEDF and MMWM datasets with a training, validation, and test set split ratio of 6:2:2. The CLP strategy was ultilized for all of the aforementioned datasets. On the TUAB and TUEV datasets, we strictly followed the settings specified in previous studies~\cite{ref16, ref22} using a fine-tuning strategy. In CLP, the dropout rate for Spectro-Spatial Projection was set to 0.1, and the dropout rate for the Linear Probe Head was set to 0.5, with a uniform learning rate of 5e-4.

All experiments were conducted on a Linux system running Ubuntu 20.04, equipped with four NVIDIA RTX 3090 GPUs, each with 24 GB of memory. All experimental results are the average of three trials after random initialization of the model.

\subsection{Downstream Results and Analysis}

We first conducted comparative experiments on the TUAB and TUEV datasets, with the results summarized in Tables~\ref{tab2:tuab_results} and~\ref{tab3:tuev_results}, respectively. On the TUAB dataset, both the Base and Deep variants of DARE-EEG consistently outperformed all competing methods. In particular, DARE-EEG-Deep achieved improvements of more than 1.6\% in balanced accuracy and 1.7\% in AUROC over the previous state-of-the-art method, EEGPT. On the TUEV dataset, DARE-EEG further demonstrated its superiority by improving balanced accuracy by over 3\% compared to the strongest baseline. These results indicate that DARE-EEG not only delivers superior performance but also exhibits strong robustness across different EEG classification tasks. Moreover, it is noteworthy that DARE-EEG-Base achieves competitive and even superior performance despite employing a substantially smaller pre-trained encoder than other comparison methods (approximately 6M vs. 20M), further validating the effectiveness of enforcing the MIP. The detailed configuration of the DARE-EEG parameters are discussed in the \textbf{\textit{Section~\ref{sec3.3.1:Para abl}}}.

We further evaluated DARE-EEG on the remaining five datasets using the Deep architecture. In addition to comparisons with existing pre-trained EEG foundation models, we also included several classic EEG decoding approaches, such as EEGNet~\cite{ref47:EEGNet}, DeepCNN~\cite{ref48:DeepCNN}, and EEG-Conformer~\cite{ref49:Conformer}. The results are summarized in Table~\ref{tab4:downstream_main}. DARE-EEG consistently achieved the highest balanced accuracy across all evaluated datasets, surpassing previous state-of-the-art methods. It improved balanced accuracy by more than 2.5\% on BCIC-2B and 3.0\% on SEEDIV, with the largest gain observed on the MMWM dataset, reaching approximately 4\%. Moreover, DARE-EEG ranked among the top-performing methods across other evaluation metrics, demonstrating overall superiority on BCIC-2A, BCIC-2B, SEEDIV, and MMWM.

It is worth noting that, on certain datasets, some classical supervised models outperformed pre-trained EEG models-for example, DeepCNN surpassed BENDR and BIOT on BCIC-2A, and EEG-Conformer outperformed LaBraM on SEEDIV. Nevertheless, DARE-EEG consistently outperformed all classical methods across these datasets. This advantage can be attributed to the proposed dual-aligned representation learning strategy, explicitly enforcing the MIP in the encoding space. which is precisely where classic models have an advantage (using data labels). Importantly, these performance gains are achieved without a substantial increase in model parameters. 
% the parameter count of DARE-EEG remains comparable to other pre-trained models and is significantly smaller than that of traditional DeepCNN and EEG-Conformer architectures. 
In addition, on the SEEDIV dataset, BENDR exhibited degraded performance due to its reliance on fixed electrode configurations, which have limited overlap with emotion-related channels. In contrast, DARE-EEG alleviates this issue through the proposed CLP and channel position embedding, further highlighting its robustness and adaptability to heterogeneous EEG datasets. For further discussion, please refer to \textbf{\textit{Appendix~\ref{asec:addires}}}.

\subsection{Ablation Study and Model Analysis}
In this section, we present ablation studies to analyze the effects of different model components and hyperparameter settings. Additional analyses, including CLP analysis (\textbf{\textit{Appendix~\ref{asec:clp}}}), the necessity of pre-training (\textbf{\textit{Appendix~\ref{asec:prew},~\ref{asec:sdata}}}), and classification and topographic maps (\textbf{\textit{Appendix~\ref{asec:vis1},~\ref{asec:vis2}}}) visualization results, are provided in the Appendix.

\subsubsection{Comparison of Different Parameters}
\label{sec3.3.1:Para abl}

We evaluated DARE-EEG under five architectural configurations with varying parameter scales on the BCIC-2B dataset, and the results are reported in Table~\ref{tab5:para_abl}. As model capacity increases, performance consistently improves, accompanied by corresponding reductions in all three loss terms. However, these gains gradually saturate beyond the Base and Deep configurations. Consequently, we consider the Base and Deep variants to offer an optimal trade-off between performance and complexity. A scaling law analysis is provided in the \textbf{\textit{Appendix~\ref{asec:scalaw}}}.

\subsubsection{Module Ablation}

\begin{figure}
    \centering
    \includegraphics[width=0.9\columnwidth]{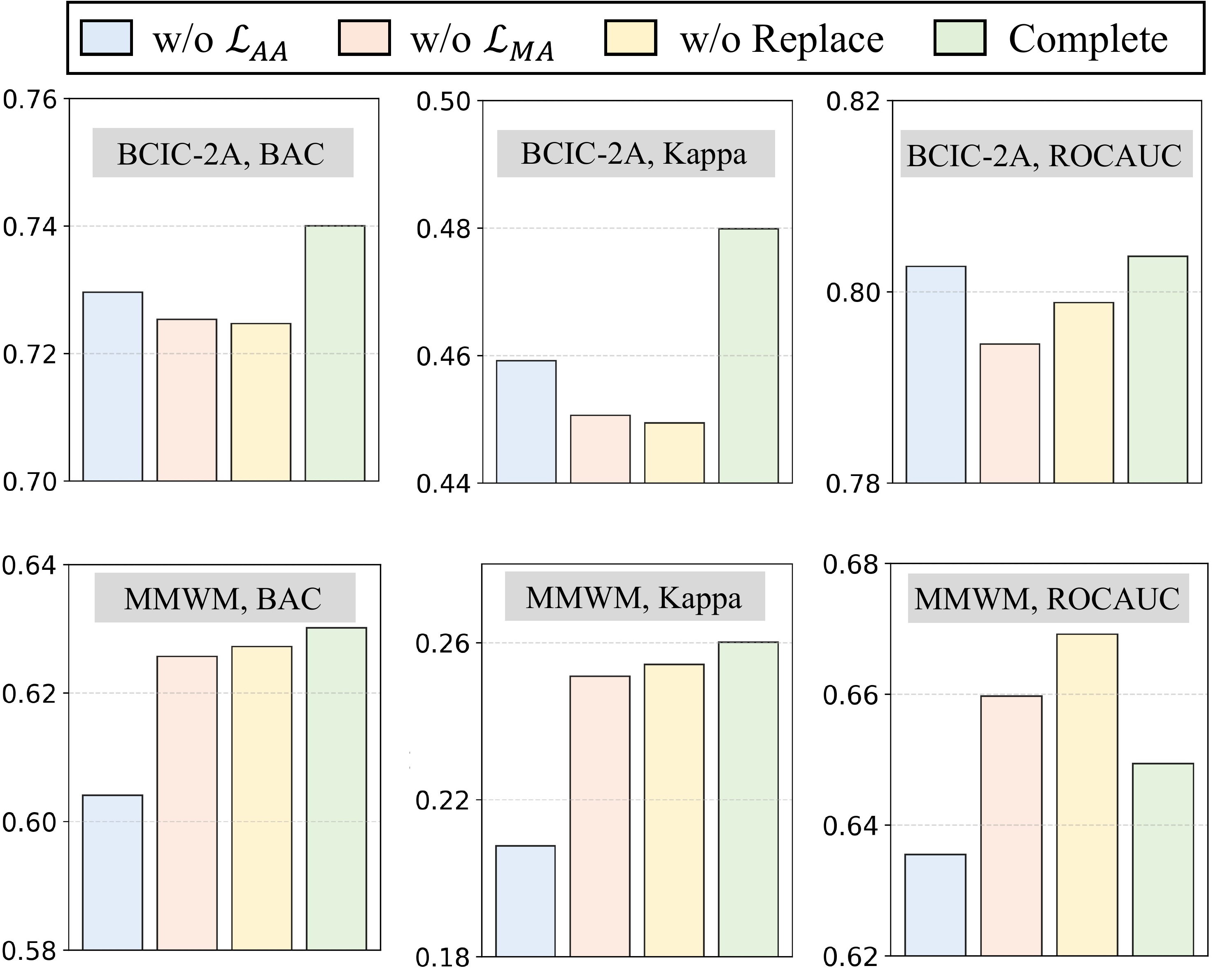}
    \caption{Results of the module ablation experiment.}
    \label{fig5:module_abla}
\end{figure}

Ablation studies were conducted on the base architecture to evaluate AA, MA, and the Replace operation. As shown in Fig.~\ref{fig5:module_abla}, removing any component consistently degrades performance. In particular, removing AA reduces balanced accuracy by about 2.6\% on MMWM, while removing MA leads to a ~1.5\% drop on BCIC-2A. These results confirm the necessity of the dual-alignment representation in DARE-EEG. Further details are shown in the \textbf{\textit{Appendix~\ref{asec:loss}}}.

\begin{table}[t]
\centering
\caption{Comparison of metrics (BAC/Kappa/ROC\_AUC) and losses (\(\mathcal{L}_{RC}\), \(\mathcal{L}_{AA}\), \(\mathcal{L}_{MA}\)) under different parameter settings. P represents trained parameters in pre-training.}
\label{tab5:para_abl}
\small
\setlength{\tabcolsep}{3pt}
\renewcommand{\arraystretch}{0.92}

\begin{tabular}{l c r c c c c c}
\toprule
\textbf{Model} & \textbf{P (M)} & $\mathbf{d_e}$ & \textbf{Layers} & \textbf{Heads} & \textbf{S.T.} & \textbf{Loss} & \textbf{Result} \\
\midrule

\multirow{3}{*}{Nano}
& \multirow{3}{*}{0.6} & \multirow{3}{*}{64}  & \multirow{3}{*}{2, 2, 3} & \multirow{3}{*}{4} & \multirow{3}{*}{1}
& 0.48 & 0.6610 $\pm$ 0.0047 \\
&&&&&& 0.23 & 0.3220 $\pm$ 0.0094 \\
&&&&&& 0.11 & 0.7214 $\pm$ 0.0088 \\
\midrule

\multirow{3}{*}{Light}
& \multirow{3}{*}{3.8} & \multirow{3}{*}{128} & \multirow{3}{*}{6, 6, 6} & \multirow{3}{*}{4} & \multirow{3}{*}{1}
& 0.45 & 0.7101 $\pm$ 0.0035 \\
&&&&&& 0.21 & 0.4203 $\pm$ 0.0071 \\
&&&&&& 0.04 & 0.7679 $\pm$ 0.0048 \\
\midrule

\multirow{3}{*}{Small}
& \multirow{3}{*}{14.7} & \multirow{3}{*}{256} & \multirow{3}{*}{6, 6, 6} & \multirow{3}{*}{4} & \multirow{3}{*}{1}
& 0.44 & 0.7392 $\pm$ 0.0045 \\
&&&&&& 0.16 & 0.4783 $\pm$ 0.0091 \\
&&&&&& 0.02 & 0.8060 $\pm$ 0.0047 \\
\midrule

\multirow{3}{*}{Base}
& \multirow{3}{*}{19.9} & \multirow{3}{*}{256} & \multirow{3}{*}{8, 8, 8} & \multirow{3}{*}{8} & \multirow{3}{*}{4}
& 0.43 & 0.7400 $\pm$ 0.0007 \\
&&&&&& 0.15 & 0.4799 $\pm$ 0.0013 \\
&&&&&& 0.02 & 0.8037 $\pm$ 0.0018 \\
\midrule

\multirow{3}{*}{Deep}
& \multirow{3}{*}{77.8} & \multirow{3}{*}{512} & \multirow{3}{*}{8, 8, 8} & \multirow{3}{*}{8} & \multirow{3}{*}{4}
& 0.42 & 0.7471 $\pm$ 0.0017 \\
&&&&&& 0.16 & 0.4941 $\pm$ 0.0033 \\
&&&&&& 0.02 & 0.8151 $\pm$ 0.0015 \\
\bottomrule
\end{tabular}
\end{table}

\section{Conclusion}

In this paper, we propose that the MIP of EEG representations in the encoding space is a crucial factor affecting the model's performance on downstream tasks. Motivated by this observation, we propose DARE-EEG, a novel pre-training foundation model that constrains masked reconstruction through the joint use of AA and MA, enabling the EEG encoder to learn Dual-Aligned REpresentations. Experimental results have demonstrated the effectiveness of DARE-EEG. This work advances the discovery and mining of rich yet subtle neural knowledge from EEG data underlying complex brain states.

%% \clearpage
%% \twocolumn
\bibliographystyle{ACM-Reference-Format}
\bibliography{samples/references}

\clearpage
\appendix
\onecolumn

\section{Notation Table}
This section lists the abbreviations and notations used in the main text and their meanings, as shown in Table 6.

\begin{table}[H]
    \centering
    \caption{Abbreviations and notations used in main text.}
    \begin{tabular}{l|l}
    \toprule
       Abbreviation/Symbols  &  Detailed Descriptions\\
    \midrule
       EEG & Electroencephalography \\
       SNR & Low Signal-to-Noise Ratio \\
       CNN & Convolutional Neural Network \\
       LLM & Large Language Model \\
       AA & Anchor Alignment \\
       MIP &  Mask-invariance Property of EEG in the encoding space \\
       MA & Mask Alignment \\
       CLP & Conv-Linear-Probing \\
       GLT & Global Learnable Tokens \\
       MU & Momentum Update \\
    \midrule
       \(X' \in \mathbb{R}^{C'\times T'}\) & the original data with \(C'\) channels and \(T'\) points\\
       \(X_{in} \in \mathbb{R}^{C\times T}\) & the input data of DARE-EEG \\
       \(\mathcal{M}\) & the mask set, where 1 indicates visible and 0 indicates masked \\
       \(g_{\theta}\) & the EEG Encoder and its parameters \(\theta\) \\
       \(g_\psi\) & the EEG Predictor and its parameters \(\psi\) \\
       \(h_{\phi}\) & the EEG Decoder (Reconstructor) and its parameters \(\phi\) \\
       \(C\) & the number of EEG electrode channels \\
       \(N\) & the number of EEG temporal patches \\
       \(D\) & the embedding dimension \\
       \(B\) & the batch size \\
       \(P=\{P_{ij}\}\in \mathbb{R}^{C\times N \times D},\quad i\in \{1,2,...,C\},j\in \{1,2,...,N\}\) & the patched \(X_{in}\)\\
       \(\hat{P}=\{\hat{P}_{ij}\}\in \mathbb{R}^{C\times N \times D},\quad i\in \{1,2,...,C\},j\in \{1,2,...,N\}\) & the reconstructed patches \\
       \(ch_{i},\quad i \in \{1,2,...,C\}\) & the channel position coding \\
       \(P_{CE}\in \mathbb{R}^{C\times N \times D} \) & the patches after channel embedding \\
       \(\hat{P}_{CE}\in \mathbb{R}^{C\times N \times D}\) & the patches after channel embedding and reconstruction \\
       \(S\) & the number of GLTs \\
       \(m\) & the number of masked parts need padding before the reconstructor \\
       \(mN\) & the number of visible EEG temporal patches \\
       \(r_{j},\quad j \in \{1,2,...,N\}\) & the RoPE coding \\
       \(Tokens_{enc}\in \mathbb{R}^{S\times m N\times D}\) & the output of EEG Encoder \\
       \({Tokens}_{pred}\in\mathbb{R}^{S\times N\times D}\) & the output of EEG Predictor \\
       \({Tokens}_{rec}\in\mathbb{R}^{C\times N}\) & the output of EEG Reconstructor \\
       \(\mathcal{L}_{\mathrm{RC}}\) & the reconstruction loss\\
       \(\mathcal{L}_{\mathrm{AA}}\) & the anchor alignment loss\\
       \(\mathcal{L}_{\mathrm{MA}}\)& the mask alignment loss \\
       \(\mathcal{L}_{\mathrm{CE}}\)& the cross-entropy loss \\
       \(f_{\Delta}\) & the Target Encoder and its parameters \(\Delta\) \\
       \(\mathrm{Tokens}^{i}_{\mathcal{M}}\in \mathbb{R}^{S\times m N\times D}, \quad i \in \{1,2,...,B\}\) & the output of the EEG Encoder for the \(i\)-th sample in a mini-batch with mask \(\mathcal{M}\)\\
    \bottomrule
    \end{tabular}
    \label{tab:placeholder}
\end{table}

\section{Additional Experimental Results}
\label{asec:addires}

\subsection{Scaling Law Analysis}
\label{asec:scalaw}

To explore the trends in balanced accuracy and total loss as the number of DARE-EEG parameters increases in Section~\ref{sec3.3.1:Para abl}, a saturation function was used to fit the observed trends. The function formula is as follows:

\begin{equation}
\label{eq:17}
    BAC(x) = A - Be^{-klog(x)}
\end{equation}

\begin{equation}
\label{eq:18}
    Loss(x) = A + Be^{-klog(x)}
\end{equation}

Figure 6 presents the fitted scaling laws for DARE-EEG. For balanced accuracy (Eq.~\ref{eq:17}), the estimated parameters are \(A=0.7666, B=0.0875, and k=0.8752\); for the total loss (Eq.~\ref{eq:18}), the fitted parameters are \(A=0.4335, B=0.1237, and k=0.9822\). Two observations can be drawn from these results. First, model performance consistently improves as the number of parameters increases. Second, extrapolation based on the fitted curves indicates that, beyond the Base and Deep configurations, an additional increase of at least 100 M parameters would be required to obtain a further 1\% improvement in performance. Consequently, the Base and Deep architectures provide a favorable trade-off between performance gains and parameter efficiency.

\begin{figure}
    \centering
    \includegraphics[width=0.9\textwidth]{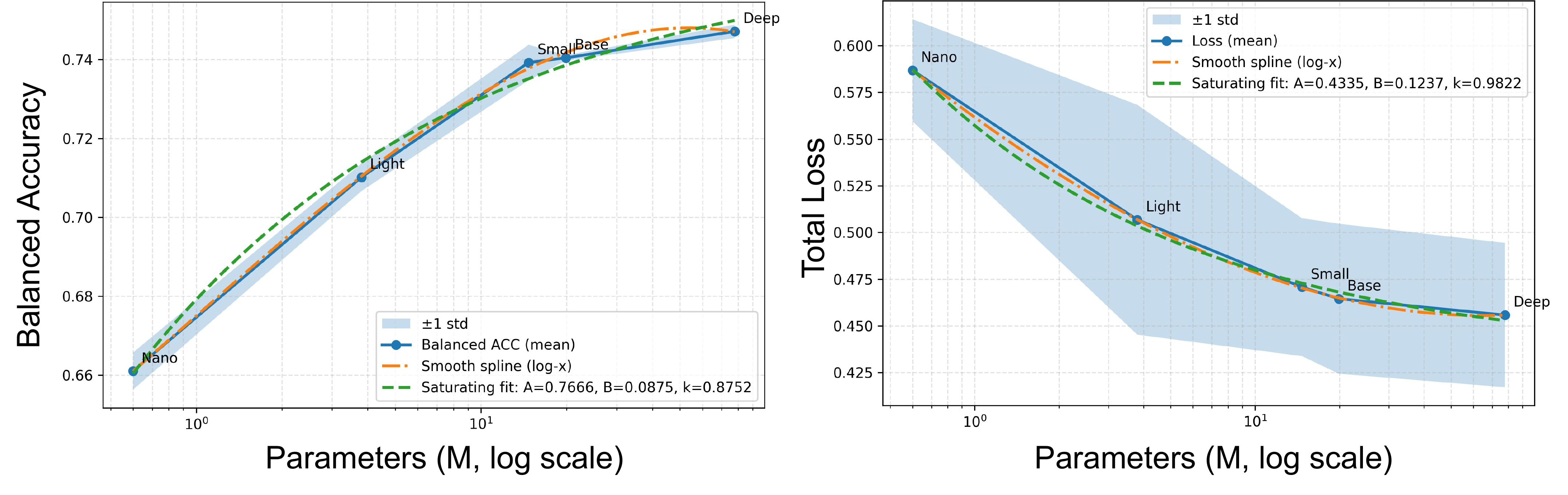}
    \caption{Scaling Laws with DARE-EEG parameters. Saturated function fitting was used. The parameter axis is on a logarithmic scale.}
    \label{fig6:ScLaw}
\end{figure}

\subsection{Loss Changes during Module Ablation}
\label{asec:loss}

\begin{figure}[H]
    \centering
    \includegraphics[width=0.9\textwidth]{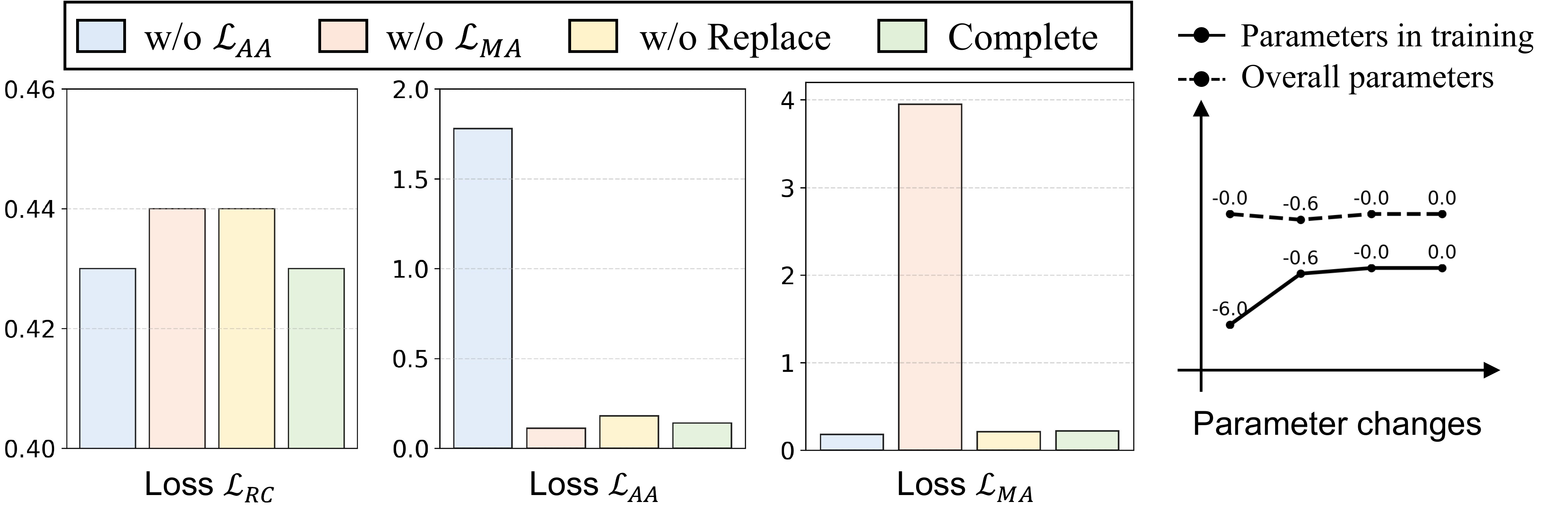}
    \caption{The changes in loss and the number of model parameters after removing different modules.}
    \label{fig7:LossAbla}
\end{figure}

Fig.~\ref{fig7:LossAbla} illustrates the evolution of the three loss terms during pre-training after selectively removing different components of DARE-EEG. The rightmost part further indicates whether the corresponding configuration leads to a reduction in model parameters. As shown, removing either AA or MA results in a dramatic increase in both AA and MA losses: the AA loss increases by up to twelve times its original value, while the MA loss rises by several-fold. This observation suggests that masked reconstruction alone is insufficient to preserve the discriminative structure of EEG representations.In addition, removing either the \(\mathcal{L}_{MA}\) or the Replace operation leads to a noticeable increase in reconstruction loss, indicating that both components play an important role in effective EEG signal reconstruction. Overall, these results demonstrate that DARE-EEG leverages MA to regularize the MIP, thereby ensuring that the representations learned by the EEG encoder remain both discriminative and robust during pre-training.

Furthermore, we observe that removing individual modules does not provide substantial reductions in model complexity. Specifically, eliminating the MA module reduces the number of trainable parameters by only approximately 0.6M, while removing the AA module has no impact on the number of trainable parameters, as the Target Encoder is updated via momentum. These results indicate that each module in DARE-EEG contributes meaningfully to performance while incurring minimal computational overhead, underscoring the effectiveness and necessity of the proposed design.

\subsection{Analysis of the CLP: Channel Projection and Training Behavior}
\label{asec:clp}

\begin{figure}
    \centering
    \includegraphics[width=0.9\textwidth]{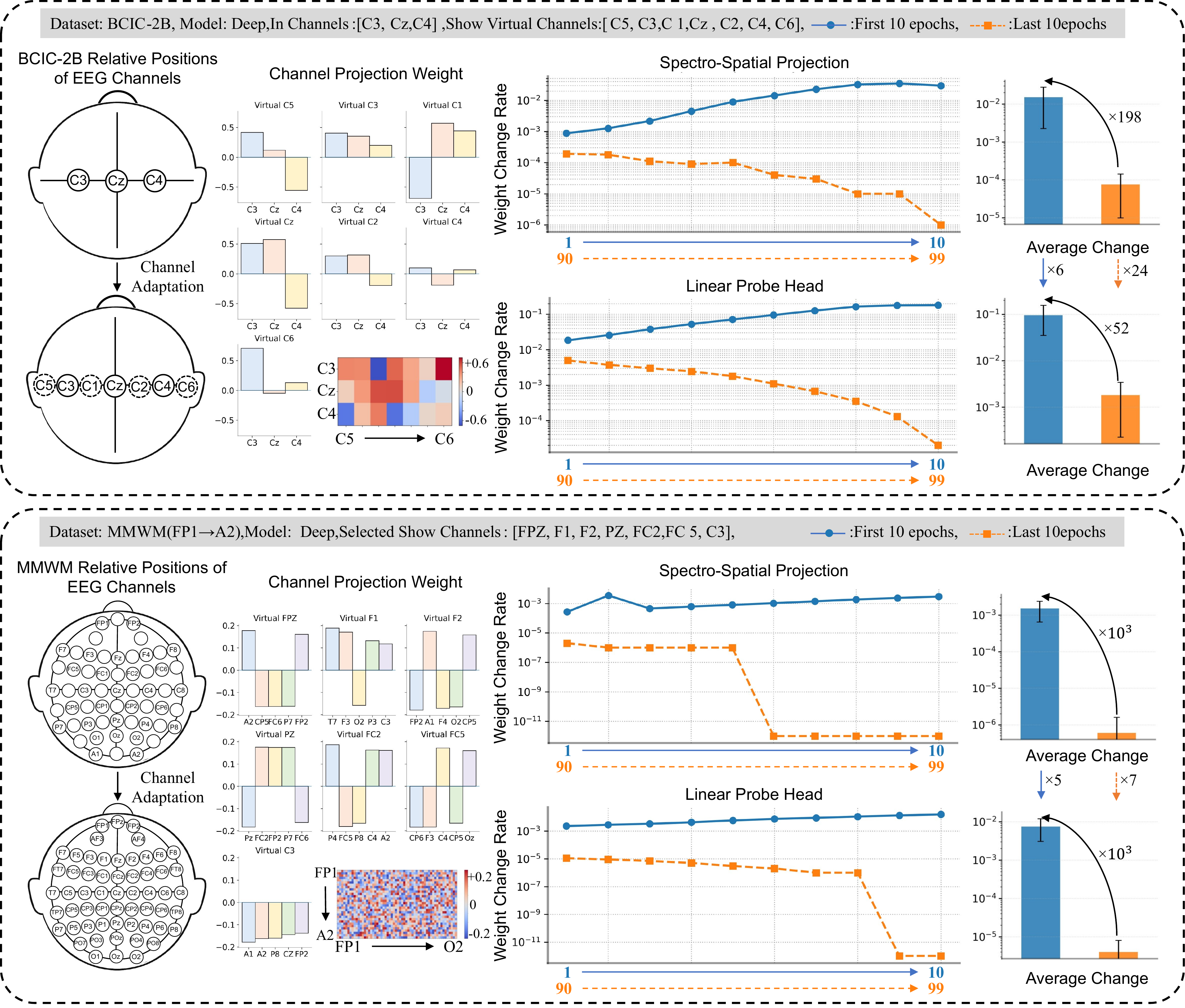}
    \caption{Analysis of the CLP during downstream training. The figure illustrates the learned channel projection weights and the parameter update magnitudes of CLP across training. Results are shown for the BCIC-2B and MMWM datasets. The original input channels and the adapted output channels after channel projection are also visualized. This analysis highlights how CLP in DARE-EEG performs channel adaptation and how its Spectro–Spatial Projection and Linear Probe Head components are progressively optimized and stabilized during training.}
    \label{fig8:CLPAnalysis}
\end{figure}

In this section, we analyze how the proposed CLP contributes to downstream adaptation in DARE-EEG. Fig.~\ref{fig8:CLPAnalysis} illustrates the learned channel projection weights on the BCIC-2B and MMWM datasets, as well as the evolution of parameter update magnitudes for the Spectro-Spatial Projection module and Linear Probe Head at the beginning and end of the training process. BCIC-2B is a binary classification dataset for motor imagery, and motor imagery belongs to a low-level human cognitive process; MMWM is a cognitive workload dataset, which belongs to a high-level human cognitive process. Using these two datasets allows us to effectively observe the differences in how DARE-EEG discovers rich representations embedded within EEG signals across different cognitive processes.

In the BCIC-2B dataset, there are only three original channels, namely [C3, Cz, C4]. We selected seven representative projected channels [C5, C3, C1, Cz, C2, C4, C6] that are strongly associated with the motor imagery task for demonstration purposes. Virtual C6 is almost entirely dominated by C3 activity (associated with right-hand motor imagery). Virtual C1/C5/Cz showed a clear contralateral contrast between C3 and C4. During training, both modules of CLP exhibited clear and consistent training dynamics. Both the Spectro-Spatial Projection and the Linear Probe Head showed significantly higher parameter weight change rates in the early stages of training compared to later stages, indicating that both modules undertook the primary adaptation and discriminative learning tasks in the initial training phase. As training progressed, the magnitude of weight updates gradually decreased and stabilized, reflecting that the model had converged to a relatively stable representation space. Further comparison reveals that the weight change rate of the Spectro-Spatial Projection was generally significantly lower than that of the Linear Probe Head, and the decrease was even more pronounced in the later stages of training. This phenomenon indicates that channel and spectral adaptation were primarily completed in the early stages of training, and this module stabilized in subsequent training phases, thus avoiding excessive perturbation to the learned representations. In contrast, the Linear Probe Head consistently maintained a relatively higher update magnitude, suggesting that it is primarily responsible for the discriminative decision-making of downstream tasks and continuously undergoes fine-grained adjustments during training.

In the MMWM dataset, The situation became more complicated. The neural mechanisms underlying working memory–induced cognitive load states are inherently complex and distributed across multiple brain regions. Moreover, datasets with a larger number of EEG channels impose higher demands on channel projection, requiring more expressive adaptation mechanisms. For clarity of analysis, we adopt a single-layer channel projection in this visualization, while the actual projection process in DARE-EEG is more sophisticated. The lower part of Fig.~\ref{fig8:CLPAnalysis} highlights a subset of channels [FPZ, F1, F2, PZ, FC2, FC5, C3] that exhibit strong correlations with cognitive workload–related tasks. The five channels with the highest contribution weights are displayed. Notably, frontal and fronto-central electrodes (e.g., FP, F, FC series) consistently receive higher absolute weights across several virtual channels, and parietal and occipital channels contribute more selectively. Unlike motor imaginary, certain electrodes of cognitive workload tasks in the original input do not necessarily remain dominant within their corresponding virtual channels after projection. Instead, the learned channel projections integrate information from multiple electrodes, reflecting the distributed and cooperative nature of neural processes underlying cognitive load. Furthermore, the training dynamics of CLP exhibit trends that are highly consistent with those observed on the BCIC-2B dataset. This consistent behavior across datasets further demonstrates the robustness of CLP, suggesting that its modular design induces stable and transferable training dynamics independent of dataset characteristics.

Overall, the distinct yet complementary training dynamics of the two CLP modules indicate that CLP in DARE-EEG achieves stable channel adaptation while retaining sufficient flexibility for task-specific discrimination. This also explains why the authors suggest setting the dropout rate for Spectro-Spatial Projection to 0.1 or even 0 in the experimental details , while setting the dropout rate for the Linear Probe Head to 0.5. This is to accommodate the characteristics of these two components: one learns a general projection matrix and stabilizes early in training, while the other continuously performs fine-grained alignment throughout the training process.

\subsection{Impact of Pre-trained Weights}
\label{asec:prew}

\begin{table}[H]
\centering
\caption{Effect of weight usage on convergence and performance. T. P. represents the parameters involved in the training, and Conv. Epochs represents the number of convergence epochs.}
\label{tab7:weight_ablation}
\small
\setlength{\tabcolsep}{4.5pt}
\renewcommand{\arraystretch}{1.0}

\begin{tabular}{c r r r r r}
\toprule
\textbf{Weight} & \textbf{T. P.} & \textbf{Conv. Epochs} & \textbf{Balanced Acc.} & \textbf{Cohen's Kappa} & \textbf{ROC\_AUC} \\
\midrule
$\times$ & 25.3\,M & 48.2 & 0.6978 & 0.3957 & 0.7560 \\
$\checkmark$ & 33.5\,K & 30.6 & 0.7471 & 0.4941 & 0.8151 \\
\bottomrule
\end{tabular}
\end{table}

In this section, we investigate the effect of using pre-trained weights on downstream tasks. Table~\ref{tab7:weight_ablation} summarizes the comparison between models trained with and without pre-trained weights. When pre-trained weights are not employed, the model is trained from scratch on the BCIC-2B dataset using supervised learning. The results indicate a substantial performance degradation without pre-training: the balanced accuracy drops by approximately 4.9\%, while the number of epochs required for convergence increases noticeably. These findings demonstrate that leveraging pre-trained weights not only improves downstream performance but also accelerates training convergence, making it clearly preferable to training models from scratch.

\subsection{Training and Evaluation with Small Dataset}
\label{asec:sdata}

\begin{table}[H]
\centering
\caption{Evaluation results on the MMWM subject-independent  small sample dataset.}
\label{tab8:smalldata}
\small
\setlength{\tabcolsep}{4.5pt}

\begin{tabular}{l c c c c}
\toprule
\textbf{Model} & \textbf{Load Weight} & \textbf{Balanced Accuracy} & \textbf{Cohen's Kappa} & \textbf{ROC\_AUC} \\
\midrule
LaBraM    & $\times$      & 0.6706 $\pm$ 0.0793 & 0.3313 $\pm$ 0.1584 & 0.6659 $\pm$ 0.1178 \\
EEGPT     & $\times$      & 0.7118 $\pm$ 0.0867 & 0.4179 $\pm$ 0.1729 & 0.7317 $\pm$ 0.1217 \\
\rowcolor{gray!20}
DARE-EEG  & $\times$      & 0.7104 $\pm$ 0.0888 & 0.4130 $\pm$ 0.1730 & 0.7284 $\pm$ 0.0925 \\
\midrule
LaBraM    & $\checkmark$  & 0.7142 $\pm$ 0.0931 & 0.4221 $\pm$ 0.1881 & 0.7395 $\pm$ 0.1107 \\
EEGPT     & $\checkmark$  & 0.7385 $\pm$ 0.0714 & 0.4711 $\pm$ 0.1415 & 0.7666 $\pm$ 0.0901 \\
\rowcolor{gray!20}
DARE-EEG  & $\checkmark$  & \textbf{0.7640 $\pm$ 0.0747} & \textbf{0.5223 $\pm$ 0.1507} & \textbf{0.7934 $\pm$ 0.0942} \\
\bottomrule
\end{tabular}
\end{table}

\begin{figure}
    \centering
    \includegraphics[width=0.9\textwidth]{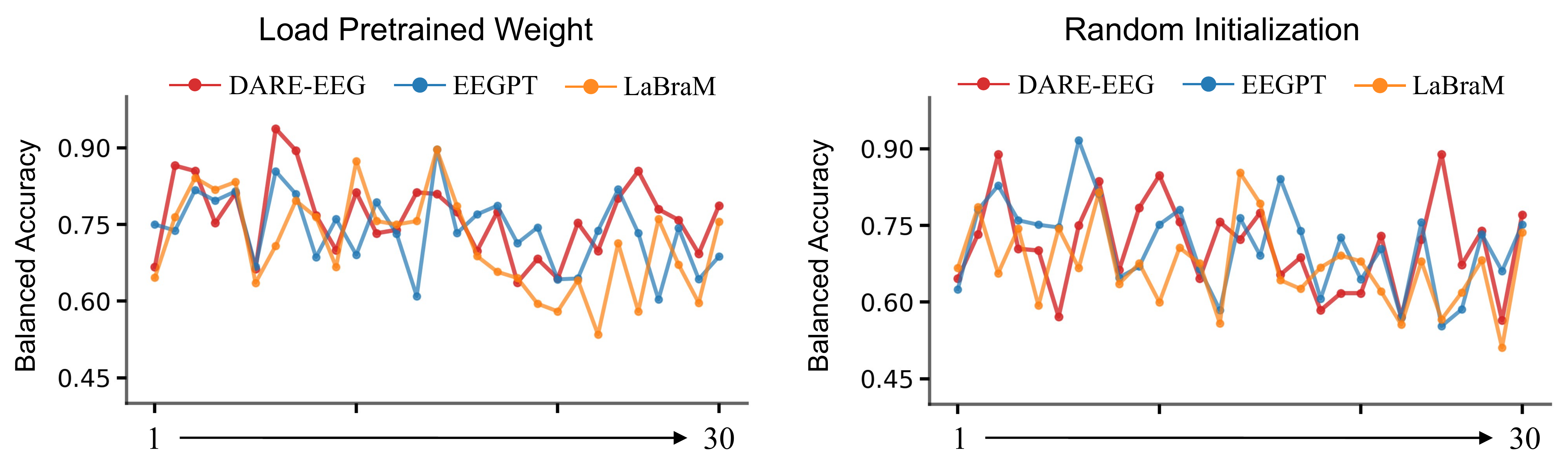}
    \caption{Small data learning performance on the MMWM dataset with subject-dependent evaluation. The line plots report results across 30 subjects, comparing models trained from random initialization with those loading pre-trained weights.}
    \label{fig9:smalldata}
\end{figure}

A key advantage of fine-tuning (or training only a lightweight classification head) a pre-trained EEG foundation model over task-specific models is its reduced tendency to overfit on limited data. To evaluate this property, we assessed the classification performance of DARE-EEG under small-data settings, with the results summarized in Table~\ref{tab8:smalldata}. On the MMWM dataset, the 29 available subjects (excluding one subject with abnormal recordings) were treated as 29 independent datasets for subject-dependent evaluation, thereby simulating a small-sample scenario. Performance statistics are reported as the mean and variance across subjects.

The results demonstrate that models initialized with pre-trained weights consistently outperform those trained from scratch. Notably, DARE-EEG achieves the best overall performance, improving balanced accuracy by approximately 2.5\% compared to the previous state-of-the-art method. In addition, Fig.~\ref{fig9:smalldata} presents detailed subject-wise evaluation results, further illustrating the robustness and consistency of the proposed approach across individual subjects.

\subsection{t-SNE Visualization Results}
\label{asec:vis1}

\subsubsection{\textbf{t-SNE Results on BCIC-2A Dataset}}

\begin{figure}[H]
    \centering
    \includegraphics[width=0.9\textwidth]{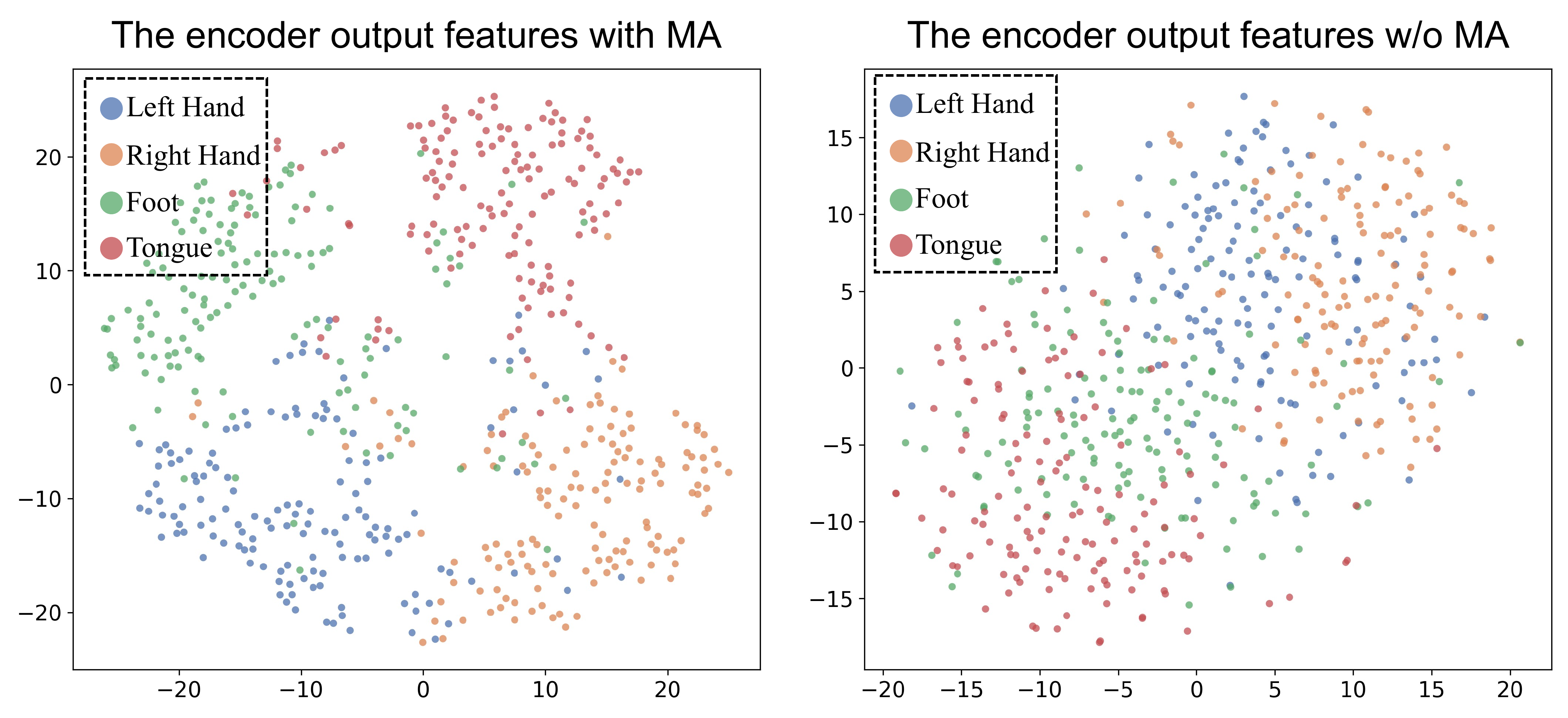}
    \caption{The t-SNE results for subject 1 on the BCIC-2A dataset are shown. The left image shows the results using MA, while the right image shows the results without using MA.}
    \label{fig10:tsne-2a}
\end{figure}

Fig.~\ref{fig10:tsne-2a} shows the t-SNE~\cite{ref51} visualization results of the output features of the EEG Encoder with and without MA on the BCIC-2A dataset. Since the EEG Encoder outputs high-dimensional EEG latent representations, we employed PCA to reduce the dimensionality of the output to 30 before applying t-SNE. It can be observed that the output of the EEG Encoder exhibits a certain degree of separability, and the classification boundaries are clearer when using MA. This indicates that MA can optimize the encoding capability of DARE-EEG to some extent.

\subsubsection{\textbf{t-SNE Results on BCIC-2B Dataset}}

\begin{figure}[H]
    \centering
    \includegraphics[width=0.9\textwidth]{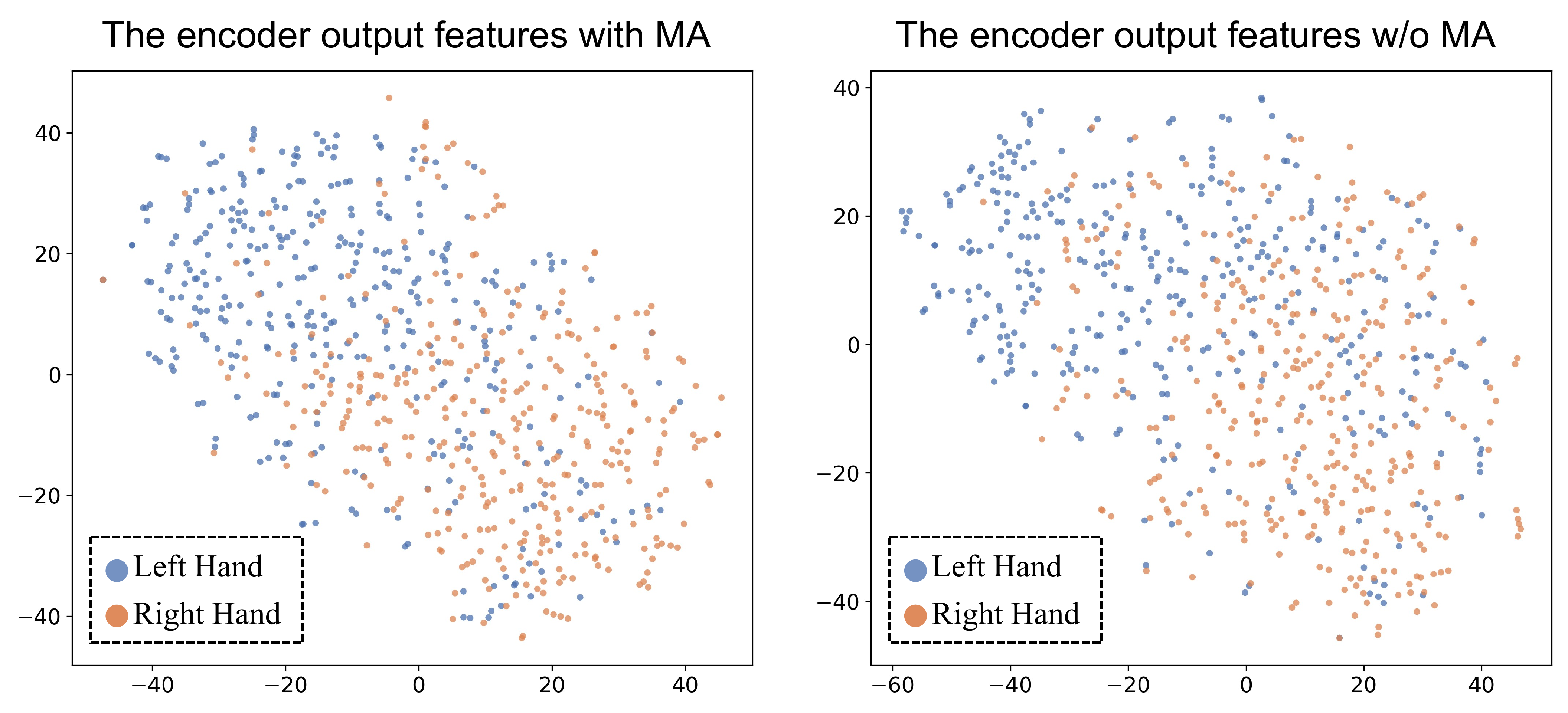}
    \caption{The t-SNE results for subject 7 on the BCIC-2B dataset are shown. The left image shows the results using MA, while the right image shows the results without using MA.}
    \label{fig11:tsne-2b}
\end{figure}

Fig.~\ref{fig11:tsne-2b} shows the t-SNE~\cite{ref51} visualization results of the output features of the EEG Encoder with and without MA on the BCIC-2B dataset. Since the EEG Encoder outputs high-dimensional EEG latent representations, we employed PCA to reduce the dimensionality of the output to 30 before applying t-SNE. It can be observed that the output of the EEG Encoder exhibits a certain degree of separability, and using MA resulted in clearer clustering properties. This indicates that MA can optimize the encoding capability of DARE-EEG to some extent.

\subsection{EEG Topography Analysis}
\label{asec:vis2}

\begin{figure}[H]
    \centering
    \includegraphics[width=0.9\textwidth]{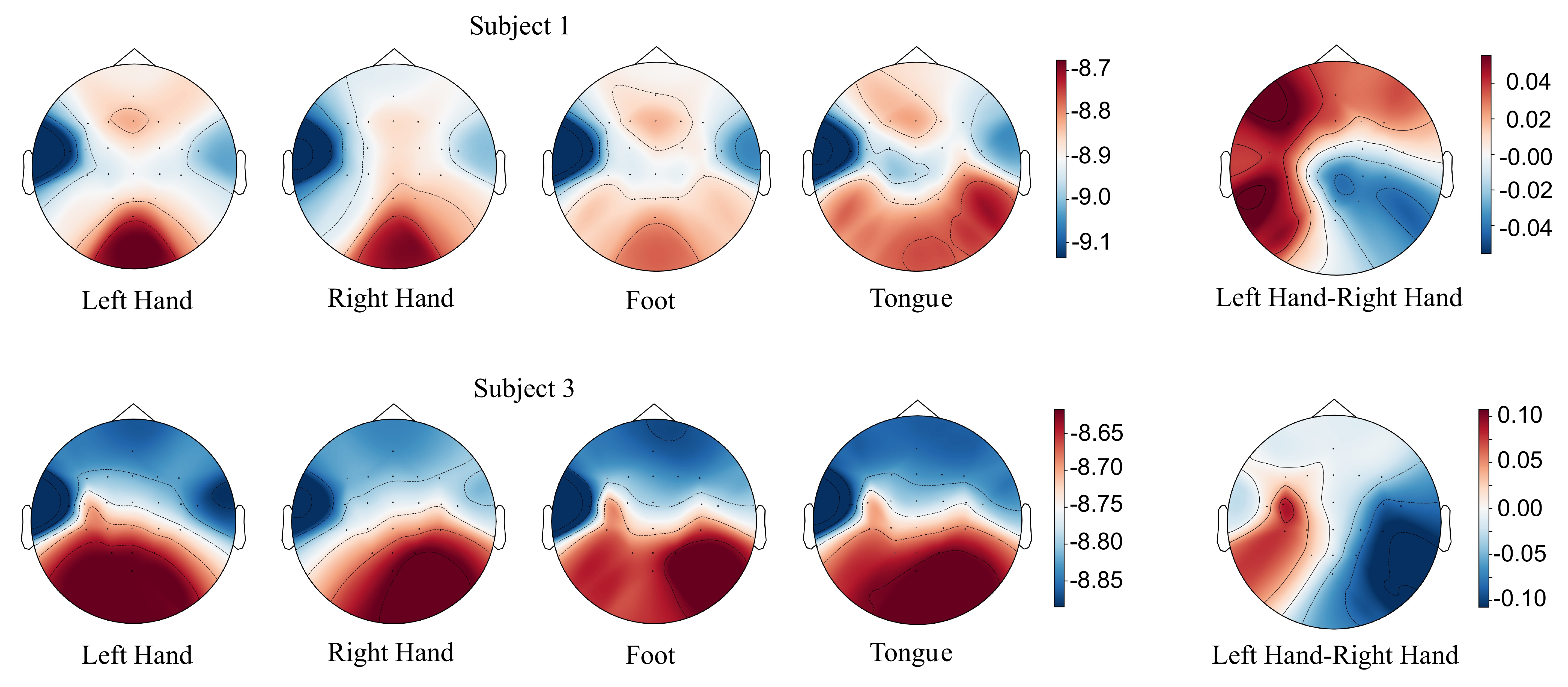}
    \caption{PSD topographic maps of Subjects 1 and 3 from the BCIC-2A dataset. PSD values are shown on a log10 scale over the 0–38~Hz frequency range.}
    \label{fig12:eegtopo-2a}
\end{figure}

\begin{figure}
    \centering
    \includegraphics[width=0.9\textwidth]{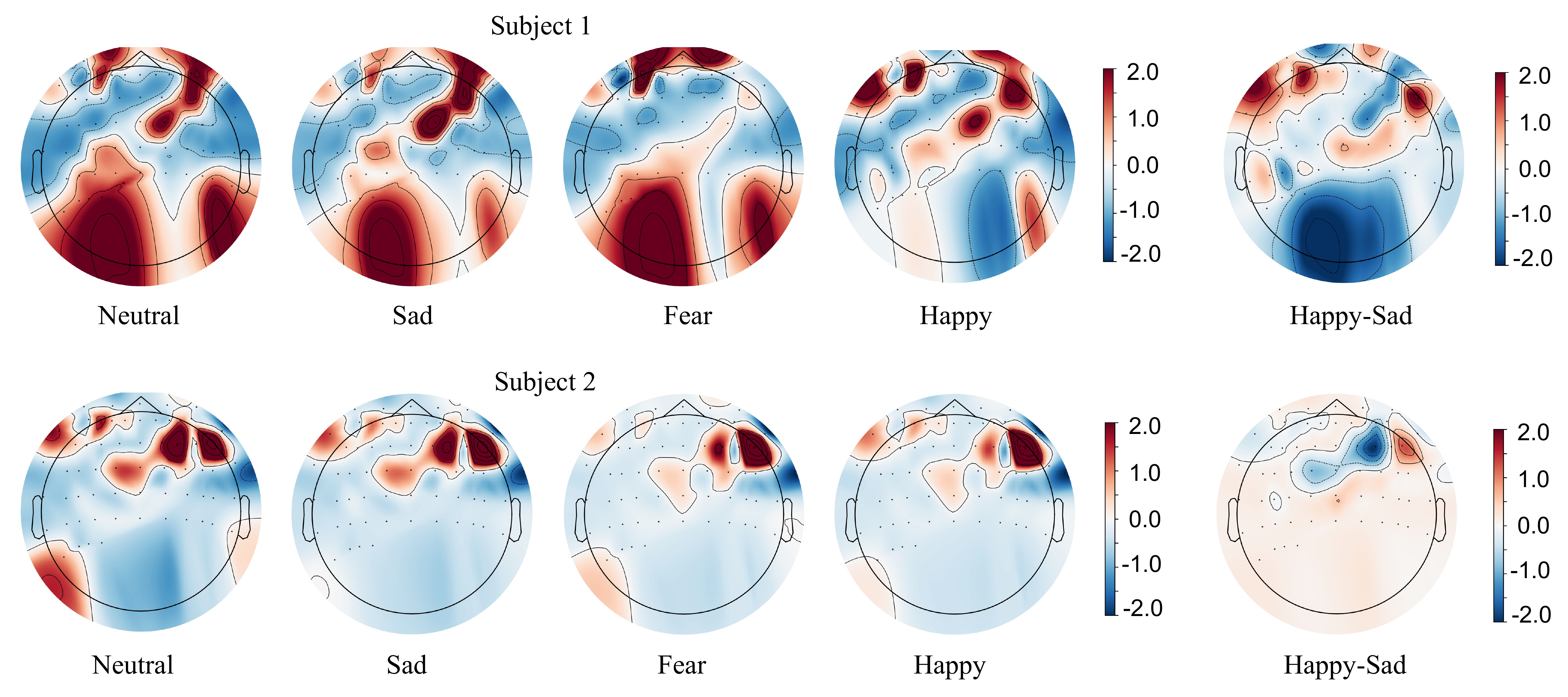}
    \caption{PSD topographic maps of Subjects 1 and 2 from the SEEDIV dataset. The gamma frequency band is shown, with values normalized using z-score normalization.}
    \label{fig13:seediv}
\end{figure}

Fig.~\ref{fig12:eegtopo-2a} and~\ref{fig13:seediv} present the Power Spectral Density (PSD) visualizations for the BCIC-2A and SEEDIV datasets, respectively. The PSD features were computed using the Welch method~\cite{ref52}, while the SEEDIV dataset provides pre-extracted and temporally smoothed PSD features.

For the motor imagery task in BCIC-2A, relatively clear and consistent spectral patterns can be observed across subjects. As illustrated by the differential PSD on the right side of Fig.~\ref{fig12:eegtopo-2a}, the contrast between left-hand and right-hand motor imagery exhibits a characteristic lateralized pattern, with higher PSD values over the left hemisphere and lower values over the right hemisphere for both subjects. In contrast, the SEEDIV emotion dataset shows more pronounced inter-subject variability. As shown in Fig.~\ref{fig13:seediv}, the differential PSD between two opposing emotional states, Happy and Sad, varies substantially across subjects. This observation highlights the inherent subject-specific nature of emotional EEG patterns and explains the increased difficulty of cross-subject emotion classification.

\section{MIP Theory Proof}
\label{asec:mip_proof}

This section uses mathematical methods to theoretically demonstrate that using MA can indeed guarantee mask invariance in the encoding space, constraining the encoder to bring together EEG samples of the same category and ensuring discriminative differences between different samples.  In contrast, simply using the complete output of the target encoder for contrastive learning (anchor alignment) does not explicitly guarantee this property.

\subsection{Definition and Objectives}

After applying patching, we obtain the token sequence \(P(X)\in\mathbb{R}^{C\times N \times d}\) from originate data \(X\in \mathbb{R}^{C\times T}\). Mask \(m\in{\{0,1\}}^{N}\), The mask operator \(\mathcal{M}(\cdot;m)\) sets the token to 0 at the masked positions, while preserving the positional information.

GLS representation of EEG Encoder \(f_{\theta}\):

\begin{equation}
\label{eq:19}
    z_{\theta}(X, m)=\operatorname{GLS}
    \left(f_{\theta}(\mathcal{M}(P(X) ; m))\right) \in \mathbb{R}^{d}, 
    \quad\|z\|=1 .
\end{equation}

The updated Target Encoder processes the complete input:

\begin{equation}
    \label{eq:20}
    \bar{z}_{\Delta}(X)=\operatorname{GLS}\left(f_{\Delta}(P(X))\right), \quad\|\bar{z}\|=1 .
\end{equation}

\subsection{Proposition 1: Masked contrastive learning explicitly enforces MIP}
\label{pp1}

Masked contrastive learning uses the GLS representations of the same sample under two independent random masks as positive pairs.

\begin{equation}
    \label{eq:21}
    \mathcal{L}_{\mathrm{MI}}=-\mathbb{E}_{X, m, m^{\prime}}
    \left[\operatorname{sim}\left(z_{\theta}(X, m), z_{\theta}\left(X, m^{\prime}\right)\right)\right].
\end{equation}

Minimizing \(\mathcal{L}_{\mathrm{MI}}\) will directly reduce the following variable:

\begin{equation}
    \label{eq:22}
    \mathbb{E}_{m, m^{\prime}}\left\|z_{\theta}(X, m)-z_{\theta}\left(X, m^{\prime}\right)\right\|^{2}.
\end{equation}

Therefore, maximizing \(\mathbb{E}\left[\operatorname{sim}\left(z(X, m), z\left(X, m^{\prime}\right)\right)\right]\) is equivalent to minimizing \(\mathbb{E}\left\|z(X, m)-z\left(X, m^{\prime}\right)\right\|^{2}\). This objective directly maximizes the similarity of the positive pairs to increase their similarity. The objective function directly includes the two mask-views, \(m\) and \(m'\), therefore this is an explicit constraint.

The proof sketch used if \(\|z\|=1\):

\begin{equation}
    \label{eq:23}
    \|a-b\|^{2}=2-2 a^{\top} b=2-2 \operatorname{sim}(a, b).
\end{equation}

Note that mask invariance alone does not prevent representation collapse; discriminability is achieved when combined with contrastive negatives or supervised contrastive objectives (Prop.~\ref{pp2}).

\subsection{Proposition 2: MIP helps with intra-class aggregation and inter-class separation}
\label{pp2}

Suppose we have three EEG samples with known label relationships \(y_{1}=y_{2}\ne y_{3}\). Note that these relationships do not necessarily correspond to three distinct class labels, but rather indicate whether samples share the same underlying condition, such as the same task or subject. If using supervised contrastive learning similar to Eq.~\ref{eq:14}, and the representation of each sample is derived from a random mask, then the loss is:

\begin{equation}
    \label{eq:24}
    \mathcal{L}_{\text {SupCon }}=-\sum_{i} \frac{1}{|\mathcal{P}(i)|} \sum_{p \in \mathcal{P}(i)} \log \frac{\exp \left(\operatorname{sim}\left(z_{i}, z_{p}\right) / \tau\right)}{\sum_{a \neq i} \exp \left(\operatorname{sim}\left(z_{i}, z_{a}\right) / \tau\right)}, \quad z_{i}=z_{0}\left(X_{i}, m_{i}\right).
\end{equation}

Where \(\mathcal{P}(i)=\{p\ne i|y_{p}=y_{i}\}\) denotes the index set of positive samples that share the same label (or underlying condition) with sample \(i\). Therefore, the above loss function will simultaneously promote intra-class aggregation \(\operatorname{sim}\left(z_{1}, z_{2}\right) \uparrow\) of similar classes \((X_{1}, X_{2})\) and inter-class separation \(\operatorname{sim}\left(z_{1}, z_{3}, etc.\right) \downarrow\) of dissimilar classes \((X_{1}, X_{3}, etc.)\). Combining this with Proposition~\ref{pp1}, we can conclude that both mask invariance and inter-class/intra-class discriminability hold simultaneously.

\subsection{Proposition 3: Anchor alignment alone cannot guarantee MIP}

The loss function for anchor alignment has already been given in Eq.~\ref{eq:11}; here we express it explicitly:

\begin{equation}
    \label{eq:25}
    \mathcal{L}_{\mathrm{AA}} = -\mathbb{E}_{X, m} [\mathrm{sim}(z_{\theta}(X, m), \bar{z}_{\Delta}(X))].
\end{equation}

Note that Eq.~\ref{eq:25} only contains \(\mathrm{sim}(z_{\theta}(X, m), \bar{z}_{\Delta}(X))\), and \(\mathrm{sim}(z_{\theta}(X, m), z_{\theta}(X, m'))\) does not appear in it. Therefore, from the perspective of the optimization objective, it does not imply that the value of \(\mathbb{E}_{m, m^{\prime}}\left\|z_{\theta}(X, m)-z_{\theta}\left(X, m^{\prime}\right)\right\|^{2}\) in Eq.~\ref{eq:22} will decrease.

The following example will further illustrate this problem with a counterexample. The set of tokens is divided into two parts using a random mask: \(U\) represents the subset of tokens that are not masked in a particular masking instance (which varies with \(m\)); \(V\) represents the subset of tokens that are masked (which also varies with m). This allows for the construction of a theoretically possible EEG Encoder representation:

\begin{equation}
    \label{eq:26}
    z_{\theta}(X, m)=\operatorname{norm}\left(\bar{z}_{\Delta}(X)+\alpha r(m)\right).
\end{equation}

Where \(r(m)\) is a vector determined solely by the mask pattern and approximately orthogonal to \(\bar{z}_{\Delta}(X)\), \(\alpha>0\). Here we find that anchor alignment can still be very small because the dominant term is still \(\bar{z}_{\Delta}(X)\), and the similarity \(\operatorname{sim}\left(z_{\theta}(X, m), \bar{z}_{\Delta}(X)\right)\) remains high, especially when \(\alpha\) is not too large. However, MIP cannot be guaranteed because \(r(m)\ne r(m')\) if \(m \ne m'\). So the \(\left\|z_{\theta}(X, m)-z_{\theta}\left(X, m^{\prime}\right)\right\|\) can become very large.

Proof sketch: Assume that $r(m) \perp \bar z(X)$ and $\|\bar z(X)\| = \|r(m)\| = 1$. 
Consider the construction in Eq.~\ref{eq:26}, where $\alpha > 0$ controls the contribution of the mask-dependent component. Then the similarity between the masked representation and the full-view target representation satisfies

\begin{equation}
\mathrm{sim}\big(z_0(X,m), \bar z(X)\big)
= \frac{1}{\sqrt{1+\alpha^2}}
\approx 1 - \frac{\alpha^2}{2},
\end{equation}

which remains high when $\alpha$ is small. Therefore, the anchor alignment loss can still be minimized. However, for two different masks $m \neq m'$, since $r(m) \neq r(m')$, the distance between the corresponding representations satisfies

\begin{equation}
\| z_0(X,m) - z_0(X,m') \|
\ \text{can be large (up to } \mathcal{O}(\alpha)\text{)},
\end{equation}

indicating that anchor alignment alone does not guarantee mask invariance.

This clearly demonstrates that simply aligning with the full target is not sufficient to rule out degenerate solutions such as "encoding the mask pattern into the representation"; in mask contrastive learning, this type of solution is explicitly penalized because it directly optimizes the \(\operatorname{sim}(z(X,m),z(X,m')\). In general, anchor alignment only requires each mask-view to resemble the full view, but does not require different mask-views to resemble each other, while mask invariance specifically enforces the latter.

\section{Dataset Description}
\label{asec:dataset}
\subsection{Datasets in Pretraining}
\subsubsection{Emotion Classification: SEED~\cite{ref26}}

\paragraph{\textbf{Dataset Overview.}}
The SEED (Shanghai Jiao Tong University Emotion EEG Dataset)\footnote{https://bcmi.sjtu.edu.cn/home/seed/seed.html} is a publicly available benchmark dataset proposed by Baoliang Lu designed for emotion recognition using EEG signals. It has been widely adopted in affective computing and EEG representation learning studies due to its relatively large subject pool and multi-session recording protocol.

\paragraph{\textbf{Experimental Paradigm.}}
In the SEED dataset, subjects were exposed to a set of carefully selected emotional video clips intended to elicit three distinct emotional states: \emph{positive}, \emph{neutral}, and \emph{negative}. Each experimental session consisted of multiple trials, where each trial corresponded to watching one video clip, followed by a short resting period.

\paragraph{\textbf{Subjects and Sessions.}}
The dataset contains EEG recordings from 15 healthy subjects. Each subject participated in three independent recording sessions conducted on different days.

\paragraph{\textbf{Preprocessing.}}
EEG signals were recorded using a 62-channel electrode cap arranged according to the international 10--20 system. The signals were originally sampled at 1000~Hz and subsequently downsampled to 200~Hz in the released dataset. A band-pass frequency filter of 0-75 Hz was applied. All recordings were referenced to the left mastoid during acquisition. In our experiment, the data was interpolated to 256~Hz and divided into 10-second samples according to the start of each trial for pre-training after channel selection.

\subsubsection{Motor Movement and Imagery Tasks: PhysioMI~\cite{ref27}}

\paragraph{\textbf{Dataset Overview.}}
PhysioMI was created by Gerwin Schalk and his colleagues at the BCI R\&D Program\footnote{https://physionet.org/content/eegmmidb/1.0.0/}. This dataset consists of over 1500 one- and two-minute EEG recordings, obtained from 109 volunteers with a 64-channel BCI2000 system.

\paragraph{\textbf{Experimental Paradigm.}}
Each subject completed 14 experimental runs, including two baseline runs (eyes open and eyes closed) and twelve task runs. The tasks involved left/right or top/bottom visual cues, prompting subjects to either execute or imagine specific movements of the fists or feet corresponding to the cue location.

\paragraph{\textbf{Preprocessing.}}
The continuous EEG data was segmented into fixed-length epochs of 6 seconds each, based on events, followed by channel selection. Finally, each 6-second segment of data was aligned to a 256 Hz sampling grid, and the amplitude units were scaled to mV.

\subsubsection{Identity Identification and Verification: M3CV~\cite{ref28}}

\paragraph{\textbf{Dataset Overview.}}
The M3CV dataset is a large-scale dataset used to study the commonalities and variations of EEG signals across subjects, time, and tasks. This dataset was first released for a brain-based identity recognition and verification task\footnote{https://aistudio.baidu.com/competition/detail/315/0/related-material}, aiming to explore a method that can reliably identify individuals using EEG signals, regardless of time or task variations.

\paragraph{\textbf{Experimental Paradigm.}}
M3CV designed a total of 6 experimental paradigms, covering resting-state, transient-state sensory, steady-state sensory, cognitive oddball (P300), motor execution, and SSVEP with selective attention. During each trial, subjects were presented with controlled visual stimuli, and their neural responses were recorded continuously.

\paragraph{\textbf{Subjects and Sessions.}}
EEG data were collected from 106 subjects, with each subject participating in two recording sessions conducted on different days. Of these participants, 95 took part in the experiment twice, with an interval of more than six days between the two experiments.

\paragraph{\textbf{Preprocessing.}}
The raw data had a sampling rate of 1000Hz and underwent TP9/TP10 re-referencing, bad channel interpolation, 0.01-200Hz bandpass filtering, ICA artifact removal, and downsampling. During pre-training, we resampled it to 256Hz and converted the amplitude units to mV.

\subsubsection{SSVEP Task: TSU~\cite{ref29}}

\paragraph{\textbf{Dataset Overview.}}
The TSU (Tsinghua University Benchmark) dataset is a publicly available steady-state visual evoked potential (SSVEP) EEG dataset collected for SSVEP-based brain--computer interface research\footnote{https://bci.med.tsinghua.edu.cn/}. It contains EEG recordings from 35 healthy subjects performing a multi-target visual stimulation task involving 40 distinct flickering stimuli.

\paragraph{\textbf{Experimental Paradigm.}}
The experiment was designed around a cue-guided SSVEP paradigm. Each subject completed six experimental blocks, and each block consisted of 40 trials corresponding to 40 visual targets presented in a randomized order. At the beginning of each trial, a visual cue indicating the target stimulus was displayed for 0.5~s, followed by a 5~s stimulation period during which all targets flickered simultaneously. Subjects were instructed to fixate on the cued target while minimizing eye blinks. A short blank interval was inserted between trials, and rest periods were provided between blocks to reduce visual fatigue. Each trial lasted 6~s in total.

\paragraph{\textbf{Subjects and Sessions.}}
EEG data were collected from 35 subjects (17 females, aged 17--34 years). Subjects were divided into two groups according to their prior experience with SSVEP-based BCIs, including an experienced group and a naive group. Each subject participated in six blocks within a single recording session.

\paragraph{\textbf{Data Acquisition.}}
EEG signals were recorded using a 64-channel Synamps2 system (Neuroscan, Inc.) with electrodes positioned according to the international 10--20 system. The data were sampled at 1000~Hz with an amplifier passband of 0.15--200~Hz. A notch filter at 50~Hz was applied during acquisition to suppress power-line interference. Electrode impedances were maintained below 10~k$\Omega$. Event triggers were synchronized with the EEG recordings.

\paragraph{\textbf{Preprocessing.}}
The continuous EEG recordings were segmented into 6~s epochs (0.5~s pre-stimulus and 5.5~s post-stimulus onset) and subsequently downsampled to 250~Hz. Each subject’s data were stored as a four-dimensional matrix with dimensions corresponding to electrode index, time points, target index, and block index. Prior to pre-training, the EEG signals were further downsampled to 256~Hz and normalized to mV scale. A global average reference was applied, and for each trial, a temporal segment from 0 to 4~s was extracted to construct the pre-training samples.

\subsubsection{MATB Task: pBCIW~\cite{ref37}}

\paragraph{\textbf{Dataset Overview.}}
pBCI is a public passive BCI dataset\footnote{https://www.neuroergonomicsconference.um.ifi.lmu.de/pbci/} that provides multiple session EEG signals for each user, and invites participants to design the best algorithm to decode cognitive load from EEG signals of new, unknown sessions, using a training dataset containing multiple sessions.

\paragraph{\textbf{Subjects and Sessions.}}
The dataset includes 15 participants (6 females; average age 25 years). Each participant performed three independent experiments, with a one-week interval between each experiment. Each experiment included a short training/warm-up phase.

\paragraph{\textbf{Experimental Paradigm.}}
Participants completed the MATB-II task, which consisted of three 5-minute modules, each with a different difficulty level (i.e., workload level), presented in a pseudo-random order. Three different workload levels were established by varying the number and complexity of subtasks (and these levels were validated through statistical analysis of subjective and objective data, including behavioral and heart rate data).

\paragraph{\textbf{Data Acquisition.}}
The data acquisition was performed using a 64 active Ag-AgCl Electrode system (ActiCap, Brain Products Gmbh) and an ActiCHamp amplifier (Brain Products, Gmbh). One electrode could not be used and one electrode was dedicated to record cardiac activity, resulting in 62 electrodes, placed according to the international 10-20 system.

\paragraph{\textbf{Preprocessing.}}
The EEG recordings were segmented into non-overlapping 2-second epochs and initially referenced to the right mastoid electrode. A high-pass FIR filter at 1~Hz was applied, followed by noisy electrode rejection based on amplitude deviation (above two standard deviations across channels) and spherical interpolation. Artifact removal was performed using second-order blind identification (SOBI) with automated independent component classification, where muscle, cardiac, and ocular components were rejected using a 95\% threshold. The cleaned signals were then low-pass filtered at 40~Hz, re-referenced using a common average reference, and downsampled to 250~Hz. For pre-training, each epoch was further temporally interpolated to 1024 time points to ensure a fixed input length.

\subsection{Datasets in Downstream Tasks}

\subsubsection{Abnormal and Event Dection: TUAB and TUEV~\cite{ref38:tuh}}

\paragraph{\textbf{Dataset Overview.}}
The TUAB (Temple University Abnormal EEG)\footnote{https://isip.piconepress.com/projects/nedc/data/tuh\_eeg/tuh\_eeg\_abnormal/} dataset is a large-scale clinical EEG dataset collected from routine hospital recordings. It is designed for binary classification between normal and abnormal EEG patterns. Due to its clinical diversity and variability, TUAB is widely used to evaluate the robustness and generalization capability of EEG representation learning models. The TUAB dataset contains a total of 2383 subjects and over 400,000 10-second samples.

The TUEV (Temple University EEG Events)\footnote{https://isip.piconepress.com/projects/nedc/data/tuh\_eeg/tuh\_eeg\_events/} dataset is a clinically annotated EEG dataset focusing on event-level abnormal patterns. It provides expert-labeled EEG segments corresponding to various pathological events, such as periodic lateralized epileptiform discharge, generalized periodic epileptiform discharge, spike and/or sharp wave discharges, artifact, and eye movement. 

\paragraph{\textbf{Preprocessing.}}
We first removed non-EEG channels such as ECG/EMG/respiration/event markers, retaining the 23 EEG channels according to the 10-20 standard, and the sampling rate was standardized to 200Hz. Then, a 0.1–75 Hz bandpass filter and a 50 Hz notch filter were applied. Finally, fixed-length time windows of 10 seconds were extracted centered around the events.

\paragraph{\textbf{Training Details.}}
During training, we used the CLP method. First, the 23-channel EEG data was projected to 58 channels consistent with the pre-trained model, and then subjected to frequency adaptation with a kernel size of 15. The batch size was set to 128, the learning rate was set to 5e-4 and a cosine scheduler was used. The warmup epochs were set to 5, and the maximum number of epochs to 50. TUAB and TUEV have already divided the data into training and validation sets. Therefore, following some previous experiments, we will use 20\% of the training set as the test set.

\subsubsection{Motor Imagery Task: BCIC-2A and BCIC-2B~\cite{ref39:bcic}}

\paragraph{\textbf{Dataset Overview.}}
The BCIC-2A and BCIC-2B datasets are two widely used benchmark motor imagery (MI) EEG datasets released as part of the BCI Competition IV\footnote{https://www.bbci.de/competition/iv/}. Both datasets are designed for evaluating MI-based brain--computer interface algorithms under controlled experimental conditions. They consist of cue-guided motor imagery tasks involving hand movements, recorded using multi-channel scalp EEG systems following standardized protocols. Due to their well-defined paradigms, subject diversity, and extensive adoption in the literature, BCIC-2A and BCIC-2B are commonly used for benchmarking EEG decoding methods and assessing cross-subject generalization performance.

BCIC-2A provides four-class (left hand, right hand, foot, tongue) motor imagery data with higher channel density (22), while BCIC-2B focuses on a binary motor imagery task (left hand, right hand) with fewer channels (C3, C4, Cz), offering complementary evaluation scenarios. The BCIC-2A dataset contains EEG recordings from 9 subjects, each participating in two recording sessions conducted on different days. The BCIC-2B dataset includes EEG data from 9 subjects with two sessions. 

\paragraph{\textbf{Preprocessing.}}
Raw EEG signals were segmented into task-related epochs according to the provided cue markers. Standard preprocessing steps were applied following common practice in motor imagery EEG studies. In addition, the EEG signals were band-pass filtered between 0 and 38~Hz and downsampled to 256~Hz. Furthermore, Euclidean Alignment~\cite{ref50} was applied to align EEG data across subjects.

\paragraph{\textbf{Training Details.}}
For the BCIC-2A dataset, channel representations were directly transformed to 58 dimensions using CLP. For BCIC-2B, the three original channels were projected onto their seven neighboring channels to enhance spatial representation. The model was optimized using AdamW with a learning rate of 5e-4, together with a OneCycle learning rate schedule. The batch size was set to 72, and training was conducted for up to 100 epochs. A leave-one-subject-out cross-validation strategy was adopted for evaluation.

\subsubsection{Emotion Classification: SEEDIV~\cite{ref40:seediv}}
\paragraph{\textbf{Dataset Overview.}}
SEEDIV is an emotion dataset designed for multi-class emotion recognition using EEG signals\footnote{https://bcmi.sjtu.edu.cn/home/seed/seed-iv.html}. It contains recordings from 15 subjects, each participating in three sessions conducted on different days. During each session, subjects watched a set of affective video clips to elicit distinct emotional responses. The dataset includes four emotion categories: \emph{neutral}, \emph{sad}, \emph{fear}, and \emph{happy}. This multi-session and multi-class design enables the evaluation of emotion-related EEG representations under varying affective states.

\paragraph{\textbf{Preprocessing.}}
The released SEEDIV dataset provides trial-wise EEG recordings aligned with video stimuli and corresponding emotion labels. In this work, an additional preprocessing pipeline was applied prior to model training. Specifically, the EEG signals were band-pass filtered between 1 and 75~Hz using a zero-phase filter. The filtered signals were then resampled from 1000~Hz to 256~Hz. A common average reference was applied by subtracting the across-channel mean at each time point. Subsequently, each trial was segmented into non-overlapping 4-second windows (1024 samples per window at 256~Hz). When the trial length was not an exact multiple of the window size, the initial residual samples were discarded to ensure non-overlapping segmentation.

\paragraph{\textbf{Training Details.}}
SEEDIV employs a 62-channel EEG configuration following the international 10–20 system, which is adaptively transformed into 58 channels via CLP during training. A leave-one-subject-out cross-validation protocol was adopted for evaluation. Model optimization was performed using AdamW in conjunction with a OneCycle learning rate schedule, with a maximum learning rate of 5e-4, a division factor of 24, and a maximum of 100 training epochs.

\subsubsection{Sleep Staging: SleepEDF~\cite{ref41:sleep}}

\paragraph{\textbf{Dataset Overview.}}
The SleepEDF dataset is a publicly available polysomnography dataset for sleep stage classification\footnote{https://physionet.org/content/sleep-edfx/1.0.0/}. It consists of overnight EEG recordings from healthy subjects, with sleep stages annotated by clinical experts according to standard scoring rules. It provides two differential channel signals, FPz-Cz and Pz-Oz. The recordings span multiple hours per subject and 197 subjects were included. The dataset defines multiple sleep stages, such as wakefulness, rapid eye movement sleep, and non-REM sleep. Following the American Academy of Sleep Medicine sleep scoring standard, sleep stages 3 and 4 were merged into a single deep sleep stage (N3), resulting in five classes: \emph{wakefulness}, \emph{N1}, \emph{N2}, \emph{N3}, and \emph{REM}. 

\paragraph{\textbf{Preprocessing.}}
SleepEDF recordings were cropped to remove 30 minutes of wake at the beginning and end of each night, converted from volts to microvolts, low-pass filtered at 30 Hz, segmented into non-overlapping 30-s windows, mapped into five sleep stages (W/N1/N2/N3/REM with stage 3 and 4 merged), and finally standardized using channel-wise z-score normalization. Since the sleepedf dataset has a sampling rate of 100~Hz, it was also interpolated and upsampled to 256~Hz.

\paragraph{\textbf{Training Details.}}
Since each Sleep-EDF sample corresponds to a 30-second EEG segment, the resulting sequence length exceeds 7,000 time points after upsampling. To handle such long sequences, a lightweight single-layer decoder was employed after the EEG Encoder to aggregate patch-level representations, which were subsequently fed into a multi-layer perceptron for classification. For channel adaptation, all electrodes surrounding the two sets of differential electrodes were incorporated, yielding a total of 21 channels. The dataset was split into training, validation, and test sets with a ratio of 6:2:2. The batch size was set to 32, and training was conducted for up to 50 epochs, while all other hyperparameters were kept consistent with those used for the previous datasets.

\subsubsection{Cognitive Workload Recognition: MMWM~\cite{ref23}}
\paragraph{\textbf{Dataset Overview.}}
MMWM is a multimodal cognitive load dataset\footnote{https://www.sciencedirect.com/science/article/abs/pii/S0893608026000389}, including EEG and fMRI data. In our experiment, only the EEG modality was used. The experimental paradigm employed a working memory task, where each participant was asked to remember characters displayed on the screen, with different cognitive loads induced by controlling the number of characters. A total of 30 participants took part in this experiment, with 14 participants participating in a second session. A 32-channel EEG cap conforming to the 10-20 system was used during data acquisition, with 30 channels actually available for use. The cognitive load levels were primarily divided into four categories, but only two categories were used in this study.

\paragraph{\textbf{Preprocessing.}}
EEG signals were preprocessed using EEGLAB, including MRI gradient artifact removal, re-referencing, band-pass filtering (0.1--80~Hz), 50~Hz notch filtering, and downsampling to 500~Hz. Physiological artifacts were removed using ICA and visual inspection. The data were segmented into 1-second epochs with baseline correction, and each epoch was temporally interpolated to 1024 time points for subsequent frequency adaptation.

\paragraph{\textbf{Training Details.}}
As the 30 available channels in MMWM provide near whole-head coverage, they were transformed into 58 channels via channel conversion during training to ensure consistency with the pre-trained model. The dataset was split into training, validation, and test sets with a ratio of 6:2:2. A batch size of 64 was used, while all other training configurations were kept consistent with those adopted for the previous datasets.

\section{Code and Pretrained Checkpoints Acquisition}
Our code is available on github \url{https://anonymous.4open.science/r/DARE-EEG-72F7} and can be run directly. The pretrained weights will also be provided in the same repository.

\end{document}